\documentclass{article}

\usepackage[final]{neurips_2021}

\usepackage[utf8]{inputenc} 
\usepackage[T1]{fontenc}    
\usepackage{hyperref}       
\usepackage{url}            
\usepackage{booktabs}       
\usepackage{amsfonts}       
\usepackage{nicefrac}       
\usepackage{microtype}      
\usepackage{xcolor}         
\usepackage[pdftex]{graphicx} 
\usepackage{amsmath}

\usepackage{geometry}
\usepackage{wrapfig}
\usepackage{lipsum}

\usepackage{enumitem}

\usepackage{tocloft}
\usepackage{color}
\usepackage{colortbl}

\newcommand{\icol}[1]{
  \begin{smallmatrix}#1\end{smallmatrix}%
}

\title{
\vspace{-6pt}
Hyper-Representations: \\
Self-Supervised Representation Learning on Neural Network Weights for Model Characteristic Prediction  
}

\author{
    Konstantin Sch\"urholt \\
    \texttt{konstantin.schuerholt@unisg.ch} \\
    AIML Lab, School of Computer Science\\
    University of St.Gallen\\
    \AND
    Dimche Kostadinov \\
    \texttt{dimche.kostadinov@unisg.ch} \\
    AIML Lab, School of Computer Science\\
    University of St.Gallen\\
    \And
    Damian Borth \\ 
    \texttt{damian.borth@unisg.ch} \\
    AIML Lab, School of Computer Science\\
    University of St.Gallen\\
}

\begin{document}

\maketitle

\vspace{-8pt}
\begin{abstract}
Self-Supervised Learning (SSL) has been shown to learn useful and information-preserving representations. 
Neural Networks (NNs) are widely applied, yet their weight space is still not fully understood.
Therefore, we propose to use SSL to learn \textit{hyper-representations} of the weights of populations of NNs.
To that end, we introduce domain specific data augmentations and an adapted attention architecture.  Our empirical evaluation demonstrates that self-supervised representation learning in this domain is able to recover diverse NN model characteristics. 
Further, we show that the proposed learned representations outperform prior work for predicting hyper-parameters, test accuracy, and generalization gap as well as transfer to out-of-distribution settings. Code and datasets are publicly available\footnote{\url{https://github.com/HSG-AIML/NeurIPS_2021-Weight_Space_Learning}}. 
\end{abstract}
\section{Introduction}
\label{Introduction}
%
This work investigates populations of Neural Network (NN) models and aims to learn representations of them using Self-Supervised Learning.
Within NN populations, not all model training is successful, i.e., some overfit and others generalize.
This may be due to the non-convexity of the loss surface during optimization \citep{goodfellow2015qualitatively}, the high dimensionality of the optimization space, or the sensitivity to hyperparameters \citep{hanin2018start}, which causes models to converge to different regions in weight space. 
What is still not yet fully understood, is how different regions in weight space are related to model characteristics. 

Previous work has made progress investigating characteristics of NN models, e.g by visualizing learned features \citep{fleet_visualizing_2014,karpathy_visualizing_2015}. 
Another line of work compares the activations of pairs of NN models \citep{raghu_svcca:_2017, morcos_insights_2018, kornblith_similarity_2019}. 
Both approaches rely on the expressiveness of the data, and are, in the latter case, limited to two models at a time. 
Other approaches predict model properties, such as accuracy, generalization gap, or hyperparameters from the margin distribution \citep{yak_towards_2019, jiang_predicting_2019}, graph topology features \citep{corneanu_computing_2020} or eigenvalue decomposition of the weight matrices \citep{martin_traditional_2019}. In a similar direction, other publications propose to investigate populations of models and to predict properties directly from their weights or weight statistics in a supervised way \citep{unterthiner_predicting_2020,eilertsen_classifying_2020}.
However, these manually designed features may not fully capture the latent model characteristics embedded in the weight space.

Therefore, our goal is to learn task-agnostic representations from populations of NN models able to reveal such characteristics.
Self-Supervised Learning (SSL) is able to reveal latent structure in complex data without the need of labels, e.g., by compressing and reconstructing data \citep{goodfellow_deep_2016,kingma_auto-encoding_2014}. 
Recently, a specific approach to SSL called contrastive learning has gained popularity \citep{misra_self-supervised_2019,chen_simple_2020,chen_improved_2020,grill_bootstrap_2020}. 
Contrastive learning leverages inherent symmetries and equivariances in the data, allows to encode inductive biases and thus structure the learned representations. 

\begin{figure*}[t]
\begin{center}
\begin{minipage}[b]{1.0\linewidth}    
{\small
\begin{tabular}{ccccc}
\text{ $ $} \text{ $ $} \text{ $ $} I. Model Zoo Generation & \text{ $ $}  \text{ $ $}  \text{ $ $} \text{ $ $} \text{ $ $} \text{ $ $} \text{ $ $} \text{ $ $} \text{ $ $} II. Representation Learning Approach & \text{ $ $}   \text{ $ $} \text{ $ $} \text{ $ $}  III. Down. Tasks
\end{tabular}
}
\end{minipage}
\centerline{\includegraphics[trim=2 2 2 2, clip, 
width=\linewidth]{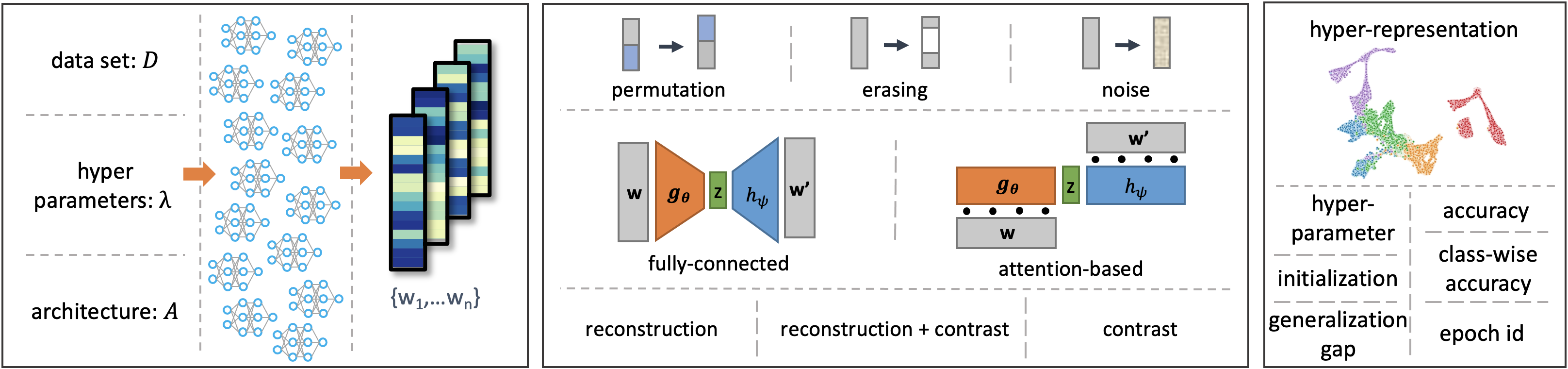}}
\caption{An overview of the proposed self-supervised representation learning approach.
        \textbf{I.} Populations of trained NNs form model zoos; each model is transformed in a vectorized form of its weights. 
        \textbf{II.} Hyper-representations are learned from the model zoos using different augmentations, architectures, and Self-Supervised Learning tasks. 
        \textbf{III.} Hyper-representations are evaluated on downstream tasks which predict model characteristics. 
        }
\label{figure.intro}
\end{center}
\vskip -0.3in
\end{figure*}

In this paper, we propose a novel approach to apply SSL to learn representations of the weights of NN populations. We learn representations using reconstruction, contrast, and a combination of both. To that end, we adapt a transformer architecture to NN weights.
Further, we propose three novel data augmentations for the domain of NN weights.
We introduce structure preserving permutations of NN weights as augmentations, which make use of the structural symmetries within the NN weights that we find necessary for learning generalizing representations. We also adapt erasing \citep{zhong2020random} and noise \citep{goodfellow_deep_2016} as augmentations for NN weights. 
We evaluate the learned representations by linear-probing for the generating factors and characteristics of the models.
An overview for our learning approach is given in Figure \ref{figure.intro}.

To validate our approach, we perform extensive numerical experiments over different populations of trained NN models. 
We find that \textit{hyper-representations}\footnote{By \textit{hyper-representation}, we refer to a learned representation from a population of NNs , i.e., a model zoo, in analogy to HyperNetworks \citep{HaDL16HyperNetworks}, which are trained to generate weights for larger NN models.} can be learned and reveal the characteristics of the model zoo.
We show that hyper-representations have high utility on tasks such as the prediction of hyper-parameters, test accuracy, and generalization gap. 
This empirically confirms our hypothesis on meaningful structures formed by NN populations. 
Furthermore, we demonstrate improved performance compared to the state-of-the-art for model characteristic prediction and outlay the advantages in out-of-distribution predictions.
For our experiments, we use publicly available NN model zoos and introduce new model zoos. 
In contrast to \citep{unterthiner_predicting_2020, eilertsen_classifying_2020}, our zoos contain models with different initialization points and diverse configurations, and include densely sampled model versions during training. 
Our ablation study confirms that the various factors for generating a population of trained NNs play a vital role in how and which properties are recoverable for trained NNs. 
In addition, the relation between generating factors and model zoo diversity reveals that seed variation for the trained NNs in the zoos is beneficial and adds another perspective when recovering NNs' properties. 


\section{Model Zoos and Augmentations}
\textbf{Model Zoo} We denote as $\mathcal{D}$ a data set that contains data samples with their corresponding labels. We denote as $\lambda$ the set of hyper-parameters used for training, (\textit{e.g.}, loss function, optimizer, learning rate, weight initialization, batch-size, epochs). We define as $A$ the specific NN architecture. Training under different prescribed configurations $\{ \mathcal{D}, \lambda, A\}$ results in a population of NNs which we refer to as \textit{model zoo}. We convert the weights and biases of all NNs of each model into a vectorized form. In the resulting model zoo $\mathcal{W} = \{ {\bf w}_1, ...., {\bf w}_M \}$, ${\bf w}_i$ denotes the flattened vector of dimension $N$, representing the weights and biases for one trained NN model. 

\label{sec:augmentations}
\textbf{Augmentations}. 
Data augmentation generally helps to learn robust models, both by increasing the number of training samples and preventing overfitting of strong features \citep{shorten_survey_2019}. 
In contrastive learning, augmentations can be used to exploit domain-specific inductive biases, e.g., know symmetries or equivariances \citep{chen_intriguing_2020}.
To the best of our knowledge, augmentations for NN weights do not yet exist. 
In order to enable self-supervised representation learning, we propose three methods to augment individual instances of our model zoos.

Neurons in dense layers can change position without changing the overall mapping of the network, if in-going and out-going connections are changed accordingly ~\citep{bishop_pattern_2006}. The relation between equivalent versions of the same network translates to permutations of incoming weights with matrix $\mathbf{P}$ and transposed permutation $\mathbf{P}^{\mathrm{T}}$ of the outgoing weights ($\mathbf{P}^{\mathrm{T}} \mathbf{P} = \mathbf{I}$).
Considering the output at layer $l+1$, with weight matrices $\mathbf{W}$, biases $\mathbf{b}$, activations $\mathbf{a}$ and activation function $\sigma$, we have 
\vspace{0.1cm}
\begin{align}
     \mathbf{z}^{l+1} &= \mathbf{W}^{l+1} \sigma(\mathbf{W}^{l} \mathbf{a}^{l-1} + \mathbf{b}^{l}) + \mathbf{b}^{l+1} = \mathbf{ \hat{W} }^{l+1} \sigma( \mathbf{ \hat{W} }^{l} \mathbf{a}^{l-1} + \mathbf{ \hat{b} }^{l}) + \mathbf{b}^{l+1},     
\end{align}
\vspace{0.1cm}
where $ \mathbf{ \hat{W} }^{l+1} = \mathbf{W}^{l+1} (\mathbf{P}^l)^{\mathrm{T}}$, $ \mathbf{ \hat{W} }^{l} = \mathbf{P}^l  \mathbf{W}^{l} $ and $ \mathbf{ \hat{b} }^{l} = \mathbf{P}^l \mathbf{b}^{l} $ are the permuted weight matrices and bias vector, respectively.
The equivalences hold not only for the forward pass, but also for the backward pass and weight update.\footnote{Details, formal statements and proofs can be found in Appendix A.} 
The permutation can be extended to kernels of convolution layers. 
The \textit{permutation augmentation} differs significantly from existing augmentation techniques. As an analogy, flips along the axis of images are similar, but specific instances from the set of possible permutations in the image domain. 
Each permutable layer with dimension $N_l$, has $N_l !$ different permutation matrices, and in total there are $\prod_l N_l!$ distinct, but equivalent versions of the same NN. 
While new data can be created by training new models, the generation is computationally expensive. The permutation augmentation, however, allows to compute valid NN samples at almost no computational cost.
Empirically, we found the permutation augmentation crucial for our learning approach.

In computer vision and natural language processing, masking parts of the input has proven to be helpful for generalization  \citep{devlin2019-bert}.
We adapt the approach of \textit{random erasing} of sections in the vectorized forms of trained NN weights. 
As in \citep{zhong2020random}, we apply the erasing augmentation with a probability $p$ to an area that is randomly chosen with a lower and upper bounds $b_{low}$ and $b_{up}$. 
In our experiments, we set $p=0.5$, $b_{low}=0.03$, $b_{up}=0.3$ and erase with zeros. Adding \textit{noise augmentation} is another way of altering the exact values of NN weights without overly affecting their mapping, and has long been used in other domains \citep{goodfellow_deep_2016}. 
%
\section{Hyper-Representation Learning}
With this work, we propose to learn representations of structures in weight space formed by populations of NNs. We evaluate the representations by predicting model characteristics.
A supervised learning approach has been demonstrated \citep{unterthiner_predicting_2020,eilertsen_classifying_2020}. \citep{unterthiner_predicting_2020} find that statistics of the weights (mean, var and quintiles) are superior to the weights to predict test accuracy, which we empirically confirm in our results. 
However, we intend to learn representations of the weights, that contain rich information beyond statistics.
‘Labels’ for NN models can be obtained relatively simply, yet they can only describe predefined characteristics of a model instance (e.g., accuracy) and so supervised learning may overfit few features, as \citep{unterthiner_predicting_2020} show. 
Self-supervised approaches, on the other hand, are designed to learn task-agnostic representations, that contain rich and diverse information and are exploitable for multiple downstream tasks \citep{lecun_self-supervised_2021}. 
Below, we present the used architectures and losses for the proposed self-supervised representation learning.

\label{sec:loss}
\textbf{Architectures and Self-Supervised Losses}. We apply variations of an encoder-decoder architecture. We denote the encoder as $g_{\theta}({\bf w}_i)$, its parameters as $\theta$, and the hyper-representation with dimension $L$ as ${\bf z}_i=g_{\theta}({\bf w}_i)$. We denote the decoder as $h_{\psi}({\bf z}_i)$, its parameters as ${\psi}$, and the reconstructed NN weights as $\hat{ {\bf w}}_i=h_{\psi}({\bf z}_i)=h_{\psi}(g_{\theta}({\bf w}_i))$. 
As is common in CL, we apply a projection head $p_{\gamma}({\bf z}_i)$, with parameters ${\gamma}$, and denote the projected embeddings as $\bar{ {\bf z}}_i=p_{\gamma}({\bf z}_i)=p_{\gamma}(g_{\theta}({\bf w}_i))$. 
In all of the architectures, we embed the hyper-representation ${\bf z}_i$ in a low dimensional space, $L<N$. 
We employ two common SSL strategies: reconstruction and contrastive learning. 
Autoencoders (AEs) with a reconstruction loss are commonly used to learn low-dimensional representations \citep{goodfellow_deep_2016}. As AEs aim to minimize the reconstruction error, the representations attempt to fully encode samples. 
Further, contrastive learning is an elegant method to leverage inductive biases of symmetries and inductive biases \citep{bronstein_geometric_2021}. 
\\
\textbf{Reconstruction (ED):} For reconstruction, we minimize the MSE $\mathcal{L}_{MSE}=\frac{1}{M}\sum_{i=1}^M \Vert {\bf w}_i - h_{\psi}({g_{\theta}({\bf w}_i)}) \Vert_2^2$.
We denote the encoder-decoder with a reconstruction loss as ED. 
\\
\textbf{Contrast (E$_c$):} For contrastive learning, we use the common NT\_Xent loss \citep{chen_simple_2020} as $\mathcal{L}_{c}$. 
For a batch of $M_B$ model weights, each sample is randomly augmented twice to form the two \emph{views} $i$ and $j$. With the cosine similarity $\text{sim}(\bar{\mathbf{z}}_i,\bar{\mathbf{z}}_j) = \bar{\mathbf{z}}_i^T\bar{\mathbf{z}}_j / ||\bar{\mathbf{z}}_i|| ||\bar{\mathbf{z}}_j||$, the loss is given as 
\begin{align}
{\mathcal{L}_{c}} = \sum_{(i,j)} \, - \log \frac{ \exp( \text{sim}(\bar{\mathbf{z}}_i, \bar{\mathbf{z}}_j)/T}{\sum_{k=1}^{2M_B} \mathbb{I}_{k \neq i} \exp( \text{sim}(\bar{\mathbf{z}}_i, \bar{\mathbf{z}}_j)/ T}, 
\end{align}
where $\mathbb{I}_{k \neq i}$ is 1 if $k \neq i$ and 0 otherwise, and $T$ is the temperature parameter.
We denote the encoder with a contrastive loss as E$_c$. 
\\
\textbf{Reconstruction + Contrast (E$_c$D):} Further we combine reconstruction and contrast via $\mathcal{L} = \beta \mathcal{L}_{MSE}+(1-\beta)\mathcal{L}_{c}$
in order to achieve good quality compression via reconstruction and well-structured representations via the contrast. We denote this architecture with its loss as E$_c$D.
\\
\textbf{Reconstruction + Positive Contrast (E${_{c\text{+}}}$D):} In contrastive learning, many methods prevented mode collapse by using negative samples. The combined loss contains a reconstruction term $\mathcal{L}_{MSE}$, which can be seen as a regularizer that prevents mode collapse. Therefore, we also experiment with replacing $\mathcal{L}_c$ in our loss with a modified contrastive term without negative samples: 
\begin{align}
{\mathcal{L}_{c\text{+}}} = \sum_{i} \, - \log \left( \exp( \text{sim}(\bar{\mathbf{z}}_i^j, \bar{\mathbf{z}}_i^k) )/T   \right) = \sum_{i} \, -  \text{sim}(\bar{\mathbf{z}}_i^j, \bar{\mathbf{z}}_i^k) + \log(T )  .
\end{align}
\citep{chen_exploring_2021,schwarzer_data-efficient_2021} explore similar simplifications without reconstruction. We denote the encoder-decoder with the loss $\mathcal{L}=\beta \mathcal{L}_{MSE}+(1 - \beta) \mathcal{L}_{c\text{+}}$ as E${_{c\text{+}}}$D. 
\begin{figure*}[t]
\begin{minipage}[t]{0.48\textwidth}
\begin{center}
\includegraphics[trim=0in 0in 0in 0in, clip, width=1.00\linewidth]{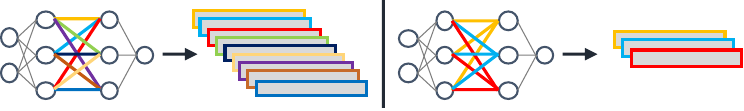}
\caption{
Trained NN weights are the input sequence to a transformer. \textbf{Left}. Each element in the sequence represents the weight that connects two neurons at two different layers. \textbf{Right}. Each element in the sequence represents the set of weights related to one neuron.}
\label{fig:transformer.input}    
\end{center}
\end{minipage}
\begin{minipage}[t]{0.02\textwidth}
\text{ }
\end{minipage}
\begin{minipage}[t]{0.48\textwidth}
\begin{center}
\includegraphics[trim=0 0 0 0, clip, width=.95\linewidth]{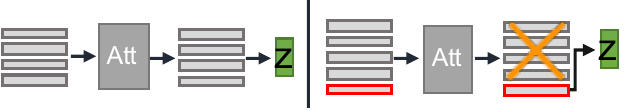}
\caption{
Multi-head attention-based encoder. 
\textbf{Left}. Regular sequence-to-sequence translation, each element of the output sequence is used.
\textbf{Right}. An additional \textit{compression token} is added to the sequence.
From the output sequence, only the compression token is taken.
}
\label{fig:compression_token}    
\end{center}
\end{minipage}
\end{figure*}
\label{sec:architectures} 

\textbf{Attention Module}. Our encoder and decoder pairs are symmetrical and of the same type. As there is no intuition on good inductive biases in the weight space, we apply fully connected feed-forward networks (FFN) as baselines. Further, wee use multi-head self-attention modules (Att) \citep{transformer_attention} as an architecture with very little inductive bias. 
In the multi-head self-attention module, we apply learned position encodings to preserve structural information \citep{Dosovitskiy2020AnII}. The explicit combination of value and position makes attention modules ideal candidates to resolve the permutation symmetries of NN weight spaces.
We propose two methods to encode the weights into a sequence (Figure \ref{fig:transformer.input}). In the first method, we encode each weight as a token in the input sequence. In the second method, we linearly transform the weights of one neuron or kernel and use it as a token. 
Further, we apply two variants to compress representations in the latent space (Figure \ref{fig:compression_token}). In the first variant, we aggregate the output sequence of the transformer and linearly compress it to a hyper-representation ${\bf z}_i$. In the second variant, similarly to \citep{devlin2019-bert,zhong2020random}, we add a learned token to the input sequence that we dub \emph{compression token}. After passing the input sequence trough the transformer, only the compression token from the output sequence is linearly compressed to a hyper-representation ${\bf z}_i$. Without the compression token, the information is distributed across the output sequence. 
In contrast, the compression token is learned as an effective query to aggregate the relevant information from the other tokens, similar to \citep{jaegle_perceiver_2021}.
The capacity of the compression token can be an information bottleneck. Its dimensionality is directly tied to the dimension of the value tokens and so its capacity affects the overall memory consumption.

\newpage
\label{proxy.about.representation.utility}
\textbf{Downstream Tasks}. We use linear probing \citep{grill_bootstrap_2020} as a proxy to evaluate the utility of the learned hyper-representations. 
As downstream tasks we use accuracy prediction (Acc), generalization gap (GGap), epoch prediction (Eph) as proxy to model versioning, F1-Score prediction (F1$_C$), learning rate (LR), $\ell_2$-regularization ($\ell_2$-reg), dropout (Drop) and training data fraction (TF). Using such targets, we solve a regression problem and measure the $R^2$ score \citep{wright1921correlation}. We also evaluate for hyper-parameters prediction tasks, like the activation function (Act), optimizer (Opt), initialization method (Init). Here, we train a linear perceptron by minimizing a cross entropy loss \citep{goodfellow_deep_2016} and measure the prediction accuracy.  

\section{Empirical Evaluation}

\subsection{Model Zoos}
\label{sec:zoos_description}

\begin{table}[t]
\begin{minipage}[t]{0.48\textwidth}
{\small
\begin{tabular}{p{.8cm}|p{.4cm}p{.4cm}p{.4cm}p{.4cm}p{.4cm}p{.4cm}p{.4cm}}
\toprule
\multicolumn{8}{c}{\texttt{TETRIS-SEED}}  \\
 & Rec. & Eph & Acc & F1$_{C0}$ & F1$_{C1}$ & F1$_{C2}$ & F1$_{C3}$ \\
\midrule
E${_{c}}$ & \text{  $ $ $ $ -} & 96.7 & \textbf{90.8} & 67.7 & 72.0 & \textbf{74.4} & 85.8 \\ 
ED & \textbf{96.1} & 88.3 & 68.9 & 47.8 & 57.2 & 33.0 & 58.1 \\
E${_{c}}$D & 84.1 & \textbf{97.0} & 90.2 & \textbf{70.7} & \textbf{75.9} & 69.4 & \textbf{86.6} \\
E${_{c\text{+}}}$D & 95.9 & 94.0 & 69.9 & 48.9 & 58.1 & 32.5 & 58.8 \\
\multicolumn{6}{c}{ } \\
\end{tabular}
}
\caption{Ablation results over self-supervised learning losses. All models implemented with attention-based reference architecture. All values are given in \%.}
\label{table.ablation:task}
\end{minipage}
\begin{minipage}[t]{0.015\textwidth}
\text{ }
\end{minipage}
\begin{minipage}[t]{0.485\textwidth}
\centering
{\small
\begin{tabular}{p{.8cm}|p{.4cm}p{.4cm}p{.4cm}p{.4cm}p{.4cm}p{.4cm}p{.4cm}}
\toprule
\multicolumn{8}{c}{\texttt{TETRIS-SEED}}  \\
 & Rec. & Eph & Acc & F1$_{C0}$ & F1$_{C1}$ & F1$_{C2}$ & F1$_{C3}$ \\
\midrule
FF          & 0.0  & 80.0 & 85.3 & 57.3 & 53.9 & 64.5 & 80.1 \\
Att$_W$     & 6.8  & 95.3 & 71.1 & 47.9 & 69.0 & 49.9 & 61.3 \\
Att$_{W+t}$ & 74.1 & 95.4 & 88.6 & 65.6 & 69.8 & \textbf{69.9} & \textbf{86.8} \\
Att$_{N}$   & \textbf{89.4} & \textbf{97}.1 & 88.4 & \textbf{71.4} & \textbf{80.8} & 69.1 & 82.3 \\
Att$_{N+t}$ & 84.1 & 97.0 & \textbf{90.2} & 70.7 & 75.9 & 69.4 & 86.6 \\
\end{tabular}
}
\caption{Ablation results in \% under E$_c$D setup. We use feed-forward FF and attention-based variants with weight and neuron encoding Att$_W$ and Att$_N$ each with $+t$ and without compress. token.}
\label{table.ablation:architecture}
\end{minipage}
\vspace{-.2in}
\end{table}

\textbf{Publicly Available Model Zoos}. \citep{unterthiner_predicting_2020} introduced model zoos of CNNs with 4970 parameters trained on MNIST \citep{mnistlecun}, Fashion-MNIST \citep{Fashion-MNIST2017}, CIFAR10 \citep{CIFAR10:DB} and SVHN \citep{Yuval:Netzer:SVHN}, and made them available under CC BY 4.0. We refer to these as \texttt{MNIST-HYP}, \texttt{FASHION-HYP}, \texttt{CIFAR10-HYP} and \texttt{SVHN-HYP}. We categorize these zoos as \emph{large} due to their number of parameters. In their model zoo creation, the CNN architecture and seed were fixed, while the activation function, initialization method, optimizer, learning rate, $\ell_2$ regularization, dropout and the train data fraction were varied between the models.

\begin{wraptable}[11]{l}{0pt}
{\small
\begin{tabular}{c|p{.5cm}|p{.4cm} p{.4cm}p{.5cm}|p{.5cm}p{.5cm}p{.5cm}|p{.585cm}}
\toprule
\multicolumn{9}{c}{\texttt{TETRIS-SEED}}  \\
&\text{  $ $ $ $ -}& P & E & N & P,E & P,N & E,N & P,E,N  \\
\midrule
ED & 88.1 & \textbf{95.5} & 89.8 & 88.8 & \textbf{96.1} & 95.6 & 89.7 & \textbf{96.1}\\
E$_c$D & 49.2 & \textbf{72.9} & 67.3 & 59.4 & \textbf{84.1} & 81.9 & 65.4 & \textbf{84.1} \\
E$_{c\text{+}}$D  & 86.3 & \textbf{95.6} & 88.5 & 87.6 & \textbf{95.9} & 95.8 & 89.0 & \textbf{96.0} \\
\end{tabular}
}
\caption{Ablation results for different augmentations and representation learning tasks. We use: permutation (P), erasing (E) and noise (N) augmentation and report test-split reconstruction $R^2$ scores in \%.
}
\label{ablation_augmentation}
\end{wraptable}
\textbf{Our Model Zoos}. We hypothesize that using only one fixed seed may limit the variation in characteristics of a zoo. To address that, we train zoos where we also vary the seed and perform an ablation study below. As a toy example to test architectures, SSL tasks and augmentations, we first create a 4x4 grey-scaled image data set that we call \emph{tetris} by using four tetris shapes. 

We introduce two zoos, which we call \texttt{TETRIS-SEED} and \texttt{TETRIS-HYP}, which we group under \emph{small}. Both zoos contain FFN with two layers and have a total number of 100 learnable parameters. In the \texttt{TETRIS-SEED} zoo, we fix all hyper-parameters and vary only the seed to cover a broad range of the weight space. The \texttt{TETRIS-SEED} zoo contains 1000 models that are trained for 75 epochs. To enrich the diversity of the models, the \texttt{TETRIS-HYP} zoo contains FFNs, which vary in activation function [\texttt{tanh}, \texttt{relu}], the initialization method [\texttt{uniform}, \texttt{normal}, \texttt{kaiming normal}, \texttt{kaiming uniform}, \texttt{xavier normal}, \texttt{xavier uniform}] and the learning rate [$1e$-$3$, $1e$-$4$, $1e$-$5$]. In addition, each combination is trained with seeds 1-100.
Out of the 3600 models in total, we have successfully trained 2900 for 75 epochs - the remainders crashed and are disregarded. Similarly to \texttt{TETRIS-SEED}, we further create zoos of CNN models with 2464 parameters, each using the MNIST and Fashion-MNIST data sets, called \texttt{MNIST-SEED} and \texttt{FASHION-SEED} and grouped them as \emph{medium}. To maximize the coverage of the weight space, we again initialize models with seeds 1-1000.
\footnote{\label{fn:details_appendix}Full details on the generation of the zoos can be found in the Appendix Section C}

\textbf{Model Zoo Generating Factors}.  
Prior work discusses the impact of random seeds on properties of model zoos. 
While \citep{yak_towards_2019} use multiple random seeds for the same hyper-parameter configuration,  \citep{unterthiner_predicting_2020} explicitly argue against that to prevent information leakage between samples. To disentangle the generating factors (seeds and hyper-parameters) and model properties, we have created five zoos with approximately the same number of CNN models trained on MNIST 
\textsuperscript{\ref{fn:details_appendix}}.
\texttt{MNIST-SEED} varies only the random seed (1-1000), \texttt{MNIST-HYP-1-FIX-SEED} varies the hyper-parameters with one fixed seed per configuration (similarly to \citep{unterthiner_predicting_2020}). 
To decouple the hyper-parameter configuration from one specific seed, \texttt{MNIST-HYP-1-RAND-SEED} draws 1 random seeds for each hyper-parameter configuration. To investigate the influence of repeated configurations, in  \texttt{MNIST-HYP-5-FIX-SEED} and \texttt{MNIST-HYP-5-RAND-SEED} for each hyper-parameter configurations we add 5 models with different seeds, either 5 fixed seeds or randomly drawn seeds.
%
\begin{figure*}[t!]
\vskip -0.0in
\begin{center}

\vskip -0.05in
\begin{minipage}[b]{1.0\linewidth}    
    \includegraphics[trim=0 0 0 0,clip, width=\linewidth]{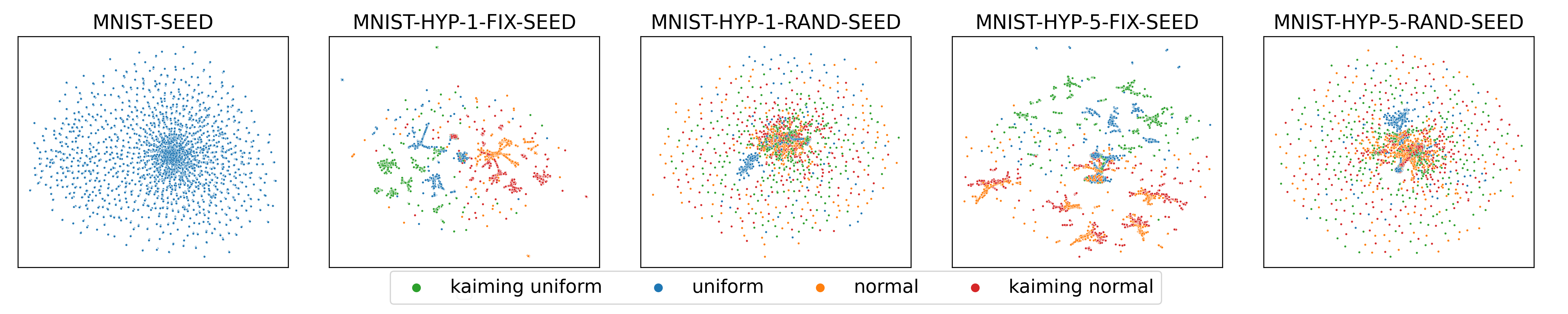}
\end{minipage}        
\vskip -0.1in
    \caption{UMAP dimensionality reduction for NN model zoos created with different generating factors. Colors represent initialization methods. Compare numerical results in Table \ref{tab:zoo_comparison}. 
    }
    \label{fig:dataset_comparison_embeddings}    
\end{center}
\vskip -0.2in
\end{figure*}
\begin{table*}[t!]
\centering
\vspace{.05in}
\begin{minipage}[b]{.99\linewidth}  
\begin{center}
\begin{small}
\begin{sc}
\hspace{-.1in}
{\small
\begin{tabular}{p{.66cm}|p{.42cm}p{.37cm}p{.52cm}|p{.42cm}p{.37cm}p{.52cm}|p{.42cm}p{.37cm}p{.52cm}|p{.4cm}p{.37cm}p{.52cm}|p{.42cm}p{.37cm}p{.52cm}}
\toprule
& \multicolumn{3}{c}{\texttt{}} & \multicolumn{3}{c}{\texttt{MNIST-HYP-} }& \multicolumn{3}{c}{\texttt{MNIST-HYP-} }&
\multicolumn{3}{c}{\texttt{MNIST-HYP-} } & \multicolumn{3}{c}{\texttt{MNIST-HYP-} } \\
& \multicolumn{3}{c}{\texttt{MNIST-SEED}} & \multicolumn{3}{c}{\texttt{1-FIX-SEED}}
& \multicolumn{3}{c}{\texttt{1-RAND-SEED}}
&
\multicolumn{3}{c}{\texttt{5-FIX-SEED}}
& \multicolumn{3}{c}{\texttt{5-RAND-SEED}}\\
\midrule
var & \multicolumn{3}{c}{.234} & \multicolumn{3}{c}{.155} & \multicolumn{3}{c}{.152} & \multicolumn{3}{c}{.091} & \multicolumn{3}{c}{.092} \\
var$_c$ & \multicolumn{3}{c}{.234} & \multicolumn{3}{c}{.101} & \multicolumn{3}{c}{.164} & \multicolumn{3}{c}{.094} & \multicolumn{3}{c}{.100}\\
\midrule
$R_{tr}$ & \multicolumn{3}{c}{73.4} & \multicolumn{3}{c}{91.3} & \multicolumn{3}{c}{80.0} & \multicolumn{3}{c}{81.3} & \multicolumn{3}{c}{65.6} \\
$R_{tes}$ & \multicolumn{3}{c}{59.0} & \multicolumn{3}{c}{72.9} & \multicolumn{3}{c}{37.2} & \multicolumn{3}{c}{70.0} & \multicolumn{3}{c}{38.6} \\
\toprule
& W & s(W) & E${_{c\text{+}}}$D & W & s(W) & E${_{c}}$D & W & s(W) & E${_{c\text{+}}}$D & W & s(W) & E${_{c}}$D  & W & s(W) & E${_{c}}$D\\
\midrule
Eph & 84.5 & \textbf{97.7}& 97.3 & 07.2 & 06.2 & \textbf{11.6} & \text{-}34 & 00.5 & \textbf{04.1} & \text{-}2.7 & 7.4 & \textbf{10.2} & 
\text{-}14 & \textbf{09.6} & 07.2 \\
Acc & 91.3 & 98.7 & \textbf{98.9} & 73.6 & 72.9 & \textbf{89.5} & \text{-}06 & 60.1 & \textbf{85.7} & 57.7 & 75.1 & \textbf{79.4} & 
\text{-}13 & 74.5 & \textbf{82.5}\\
GGap & 56.9 & 66.2 & \textbf{66.7} & 55.3  & 42.1 & \textbf{61.9} & \text{-}36 & 31.1 & \textbf{54.9} & 37.8 & 36.7 & \textbf{57.7} &
\text{-}3.2 & 46.1 & \textbf{60.5} \\
Init &  -- & -- & -- & \textbf{87.7} & 63.3 & 75.4 & 38.3 & \textbf{56.1} & 48.2 & \textbf{80.6} & 54.5 & 51.4 & 
35.5 & \textbf{47.2} & 40.4 \\
\end{tabular}
}
\end{sc}
\end{small}
\end{center}
\end{minipage}
\caption{$R^2$ score in \%. Results on the impact of the generating factors for the model zoos. \textbf{Top:} VAR is the variance of the weights. VAR$_c$ denotes the mean of the variances for groups of samples with shared initialization method and activation function. $R_{tr}$ and $R_{tes}$ are the reconstruction $R^2$ of train and test split on a reference E$_{c\text{+}}$D architecture trained for 500 epochs. \textbf{Bottom:} Downstream task performance compared to baselines predicting epochs, accuracy, generalization gap and initialization.}
\label{tab:zoo_comparison}
\end{table*}

\subsection{Training and Testing Setup}
\textbf{Architectures}. We evaluate our approach with different types of architectures, including E${_c}$, ED, E${_c}$D and E${_{c\text{+}}}$D as detailed in Section \ref{sec:loss}. 
The encoder E and decoders D in the FFN baseline are symmetrical 10 [FC-ReLU]-layers each and linearly reduce dimensionality for the input to the latent space $\mathbf{z}$. Considering the attention-based encoder and decoder, on the \texttt{TETRIS-SEED} and \texttt{TETRIS-HYP} zoos, we used 2 attention blocks with 1 attention head each, token dimensions of 128 and FC layers in the attention module of dimension 512. On the larger zoos, we use up to 4 attention heads in 4 attention blocks, token dimensions of up to size 800 and FC layers in the attention module of dimension 1000. For the combined losses, we evaluate different $\beta \in [0.05,0.85]$.

\textbf{Hyper-Representation Learning and Downstream Tasks}. We apply the proposed data augmentation methods for representation learning (see Section \ref{sec:augmentations}). We run our representation learning algorithms for up to 2500 epochs, using the adam optimizer \citep{Kingma:2014:Adam}, a learning rate of 1e-4, weight decay of 1e-9, dropout of 0.1 percent and batch-sizes of 500. In all of our experiments, we use 70\% of the model zoos for training, 15\% for validation and 15\% for testing. We use checkpoints of all epochs, but ensure that samples from the same models are either in the train, validation or the test split of the zoo. As quality metric for the hyper-representation learning, we track the reconstruction $R^2$ on the validation split of the zoo and report the reconstruction $R^2$ on the test split. 
As a proxy for how much useful information is contained in the hyper-representation, we evaluate on downstream tasks as described in Section \ref{proxy.about.representation.utility}. To ensure numerical stability of the solution to the linear probing, we apply Tikhonov regularization \citep{tikhonov1977solutions} 
with regularization parameter ${\alpha}$ in the range [$1e$-$5$, $1e3$] (we choose ${\alpha}$ by cross-validating over the $R^2$ score of the validation split of the zoo) and report the $R^2$ score of the test split of the zoo. To minimize the cross entropy loss for the categorical hyper-parameter prediction, we use the adam optimizer with learning rate of $1e$-$4$ and weight-decay of $1e$-$6$. The linear probing is applied to the same train-validation-test splits as it is in our representation learning setup.  

\textbf{Out-of-Distribution Experiments}. We follow a setup for out-of-distribution experiments similar to \citep{unterthiner_predicting_2020}. We investigate how well the linear probing estimator computed over hyper-representations generalize to yet unseen data. 
Therefore, we use the zoos \texttt{MNIST-HYP}, \texttt{FASHION-HYP}, \texttt{CIFAR10-HYP} and \texttt{SVHN-HYP}. 
On each zoo, we apply our self-supervised approach to learn their corresponding hyper-representations and fit a linear probing estimator to each of them (in-distribution).
We then apply both the hyper-representation mapper and the linear probing estimator of one zoo 
on the weights of the other zoos (out-of-distribution). 
The target ranges and distributions vary between the zoos. Linear probe prediction may preserve the relation between predictions, but include a bias. Therefore, we use Kendall’s $\tau$ coefficient as performance metric, which is a measure of rank correlation. 
It measures the ordinal association between two measured quantities
\citep{kendall:1938:measure:Biometrika}. 
\begin{table}[t]
\vspace{-.05in}
\centering
\begin{center}
{ \small
\begin{minipage}[b]{.49\linewidth}
\begin{tabular}{l|p{.4cm}p{.4cm}p{.4cm}p{.4cm}p{.4cm}p{.4cm}p{.5cm}|}
\toprule
& \multicolumn{7}{c|}{\texttt{TETRIS-SEED}}  \\
 & W & PCA$_l$ & PCA$_c$ & PCA$_{\text{r}}$ & U$_{m}$ & s(W) & E${_{c}}$D \\
\midrule
Eph & 55.4 & 46.9 & 77.8 & 93.5 & 0.00 & 96.4 & \textbf{97.0} \\
Acc & 49.1 & 23.9 & 74.7 & 74.9 & 0.01 & 86.9 & \textbf{90.2}  \\
Ggap & 43.9 & 28.3 & 72.9 & 75.7 & 0.01 & 81.0 & \textbf{81.9}  \\
\midrule
$F_{C0}$ & 26.6 & 0.07 & 52.8 & 49.1 & 0.02 & 62.5 & \textbf{70.7}  \\
$F_{C1}$ & 41.4 & 30.2 & 48.6 & 58.6 & 0.01 & 63.1 &  \textbf{75.9} \\
$F_{C2}$  & 41.5 & 0.06 & 38.3 & 38.5 & 0.01 & 60.9 & \textbf{69.4}  \\
$F_{C3}$  & 53.9 & 44.6 & 73.8 & 72.2 & 0.00 & 75.2 & \textbf{86.6} 
\end{tabular}
\end{minipage}
\begin{minipage}[b]{.495\linewidth}
\centering
\begin{tabular}{l|p{.4cm}p{.4cm}p{.4cm}p{.4cm}p{.4cm}p{.4cm}p{.4cm}}
\toprule
& \multicolumn{7}{c}{\texttt{TETRIS-HYP}}  \\
& W & PCA$_l$ & PCA$_c$ & PCA$_{\text{r}}$ & U$_{m}$ & s(W) & E${_{c}}$D \\
\midrule
Eph & 02.8 & 02.7 &  0.04 &  13.1 & 0.03 & 16.3 & \textbf{16.7}  \\
Acc & 11.0 & 14.0 & 0.81 & 73.8 & 0.7 & 80.2 & \textbf{83.7}    \\
GGap & 12.9 & 12.9 & 0.31 & 76.0 & 0.3 & 79.2 & \textbf{81.6}    \\
\midrule
LR & \text{-}4.7 & \text{-}3.2 & \text{-}1.3 & \text{-}0.4 &  \text{-}1.4 & \textbf{53.4} & 51.1  \\
Act & 74.7 & 72.7 & 45.1 &  74.2 & 45.9 &  73.6 & \textbf{86.9}   \\
Init & 38.5 & 36.4 & 32.7 & 35.4 & 33.6 & 46.6 & \textbf{47.1} \\
\multicolumn{8}{c}{ } \\
\end{tabular}
\end{minipage}
}
\end{center}
\caption{$R^2$ given in \%. \textbf{Left}.  Reconstruction, epoch, accuracy, generalization gap and class-wise F-scores prediction. \textbf{Right}. Epoch, accuracy, generalization gap, learning rate, seed, activation function and initialization prediction. 
}
\label{table.tetris-seed.and.tetris-hyp}
\end{table}

\textbf{Baselines, Computing Infrastructure and Run Time}. 
As baseline, we use the model weights (W). In addition, we also consider PCA with linear (PCA$_l$), cosine (PCA$_c$), and radial basis kernel (PCA$_{\text{r}}$), as well as UMAP (U$_{\text{m}}$) \citep{mcinnes_umap_2018}. We further compare to layer-wise weight-statistics (mean, var, quintiles) s(W) as in \citep{unterthiner_predicting_2020,eilertsen_classifying_2020}. 
As computing hardware, we use half of the available resources from NVIDIA DGX2 station with 3.3GHz CPU and 1.5TB RAM memory, that has a total of 16 1.75GHz GPUs, each with 32GB memory. To create one small and medium zoo, it takes 1 to 2 days and 10 to 12 days, respectively. For one experiment over the small zoo it takes around 3 hours to learn the hyper-representation on a single GPU and evaluate on the downstream tasks. It takes approximately 1 day for the medium zoos and 2 to 3 days for the large scale zoos for the same experiment. 
The representation learning model size ranges from ~225k on the small zoos to 65M parameters on the largest zoos.
We use ray.tune \citep{liaw2018tune} for hyperparameter optimization and track experiments with W\&B \citep{wandb}.
\begin{table*}[t]
\vspace{-.1in}
\centering
\vspace{.05in}
\begin{minipage}[b]{.99\linewidth}  
\begin{center}
\begin{small}
\begin{sc}
\hspace{-.1in}
{\small
\begin{tabular}{l|p{.6cm}p{.6cm}p{.6cm}|p{.6cm}p{.6cm}p{.6cm}|p{.6cm}p{.6cm}p{.6cm}|p{.6cm}p{.6cm}p{.6cm}}
\toprule
& \multicolumn{3}{c}{\texttt{MNIST-HYP}} & \multicolumn{3}{c}{\texttt{FASHION-HYP}} & \multicolumn{3}{c}{\texttt{CIFAR10-HYP}} & \multicolumn{3}{c}{\texttt{SVHN-HYP}} \\
& W & s(W) & E${_{c}}$D & W & s(W) & E${_{c}}$D & W & s(W) & E${_{c}}$D & W & s(W) & E${_{c}}$D   \\

\midrule
Eph & 25.8 & 33.2 & \textbf{50.0} & 26.6 & 34.6 & \textbf{51.3} & 25.7 & 30.3 & \textbf{53.3} & 
22.8 & 37.8 & \textbf{52.6} \\
Acc & 74.7 & 81.5 & \textbf{94.9} & 70.9 & 78.5 & \textbf{96.2} & 76.4 & 82.9 & \textbf{92.7} & 
80.5 & 82.1 & \textbf{91.1} \\
\midrule
GGap & 23.4 & 24.4 & \textbf{27.4} & 48.1 & 41.1 & \textbf{49.0} & 37.7 & 37.4 & \textbf{40.4} & 
38.7 & 42.2 & \textbf{44.2} \\
\midrule
LR & 29.3 & 34.3 & \textbf{37.1} & 33.5 & 35.6 & \textbf{42.4} & 27.4 & 32.3 &\textbf{44.7}  & 
24.5 & 33.4 & \textbf{49.1} \\
$\ell_2$-reg & 12.5 & 16.5 & \textbf{20.1} & 11.9 & 16.3 & \textbf{25.0} & 08.7 & 13.8 & \textbf{28.0}  &
09.0 & 13.6 & \textbf{28.0} \\
Drop & 28.5 & 19.2 & \textbf{35.8} & 26.7 & 21.3 & \textbf{38.3} & 16.7 & 16.5 & \textbf{33.8}  & 
09.0 & 14.6 & \textbf{23.3} \\
TF & 03.8 & 07.8 & \textbf{15.9} & 08.1 & 08.2 & \textbf{22.1} & 08.4 & 06.9 & \textbf{35.4}  &  
03.2 & 08.8 &  \textbf{21.4} \\
\midrule
Act & 88.6 & 81.1 & \textbf{88.7} & 89.8 & 82.4 & \textbf{90.1} & 88.3 & 80.3 & \textbf{90.0} & 
86.9 & 78.8 & \textbf{87.2} \\
Init & \textbf{94.6} & 72.0 & 80.6 & \textbf{95.7} & 76.5 & 86.7 & \textbf{93.5} & 73.3 & 82.6  & 
\textbf{91.0} & 73.0 & 82.8 \\
Opt & \textbf{76.7} & 65.4 & 66.4 & \textbf{79.9} & 67.4 & 73.0 & \textbf{74.0} & 65.5 & 71.0 & 
\textbf{72.5} & 68.2 & 72.3 \\

\end{tabular}
}
\end{sc}
\end{small}
\end{center}
\end{minipage}
\caption{\textbf{Top 7 Rows}. $R^2$ score in $\%$ for Eph, Acc, GGap, LR $\ell_2$-reg, Drop and TF prediction. \textbf{Bottom 3 Rows}. Accuracy score for Act, Init and Opt prediction. 
}
\label{comparisament.ref}
\vspace{-.15in}
\end{table*}
\subsection{Results}

\textbf{Augmentation Ablation}. 
To evaluate the impact of the proposed augmentations (Section \ref{sec:augmentations}) for our representation learning method, we present an ablation analysis on the \texttt{TETRIS-SEED} zoo, in which we measure $R^2$ for ED, E$_c$D and E$_{c\text{+}}$D\footnote{We leave out E$_c$ as it does not use a reconstruction loss}. We use 120 permutations, a probability of $0.5$ for erasing the weights, and zero-mean noise with standard deviation $0.05$ (see Table \ref{ablation_augmentation}). We find the permutation augmentation to be necessary for generalization - particularly under higher compression ratios. The additional samples generated with the permutation appear to effectively prevent overfitting of the training set. Without the permutation augmentation, the test performance diverges after few training epochs.  
Erasing further improves test performance and allows for extended training without overfitting. The addition of noise yields inconsistent results and is difficult to tune, so, we omit it in our further experiments.

\textbf{Architecture Ablation}. 
The different architectures are compared in Table \ref{table.ablation:task}. The results show that within the set of used NN architectures for hyper-representation learning, the attention-based architectures learn considerably faster, 
yield lower reconstruction error and have the highest performance on the downstream tasks compared to the FFN-based architectures. We attribute this to the attention modules, which are able to reliably capture long-range relations on complex data due the global field of visibility in each layer.  
While tokenizing each weight individually (Att$_{W}$) is able to learn, the computational load is significant, even for a small zoo, due to the large number of tokens in the sequences. The memory load prevents the application of that encoding on larger zoos. 
We find Att$_{N+t}$ embedding all weights of one neuron (or convolutional kernel) to one token in combination with compression tokens shows the overall best performance and scales to larger architectures. Compression tokens achieve higher performance, which we confirm on larger and more complex datasets. The dedicated token gathers information from all other tokens of the sequence in several attention layers. This appears to enable the hyper-representation to grasp more relevant information than linearly compressing the entire sequence. On the other hand, compression tokens are only an advantage, if their capacity is high enough, in particular higher than the bottleneck.

\textbf{Self-Supervised Learning Ablation}. In Table \ref{table.ablation:architecture}, we evaluate the usefulness of the self-supervised learning tasks (Section \ref{sec:loss}). The application of purely contrastive loss in E$_c$, learns very useful representations for the downstream tasks. However, E$_c$ heavily depends on expressive projection heads and by design cannot reconstruct samples to further investigate the representation. 
Pure reconstruction ED results in embeddings with a low reconstruction loss, but comparably low performance on downstream tasks. Among our losses, the combination of reconstruction with contrastive loss as in E$_c$D, provides hyper-representations ${\bf z}$ that have the best overall performance. The variation E$_{c+}$D, is closer to ED in general performance, but still outperforms it on the downstream tasks.  The addition of a contrastive loss with projection head helps to pronounce distinctiveness, so that the hyper-representations are good at reconstructing the NN weights, and in revealing properties of the NNs through the encoder. 
Empirically, we found that on some of the larger zoos, E$_{c+}$D performed better than E$_{c}$D. On these zoos, it appears that E$_{c+}$D is most suitable for homogeneous zoos without distinct clusters, while E$_{c}$D is suitable for zoos with more sub-structures, compare Figure \ref{fig:dataset_comparison_embeddings} and Table \ref{tab:zoo_comparison}. We therefore applied both learning strategies on all zoos and report the more performant one.

\textbf{Zoo Generating Factors Ablation}. 
Figure \ref{fig:dataset_comparison_embeddings} visualizes the weights of the zoos, which contain models of all epochs\footnote{Further visualizations can be found in the Appendix Section C}. In Table \ref{tab:zoo_comparison}, we report numerical properties. 
Only changing the seeds appears to result in homogeneous development with very high correlation between $s(W)$ and the properties of the samples in the zoo, as previous work already indicated \citep{schuerholt2021investigation}.
Varying the hyper-parameters reduces the correlation. With fixed seeds, we observe clusters of models with shared initialization method and activation function. Quantitatively we obtain lower VAR$_c$ and high predictive value of $W$ for the initialization method. 
That seems a plausible outcome, given that the architecture and activation function determines the shape of the loss surface, while the seed and initialization method decide the starting point.  
Such clustering appear to facilitate the prediction of categorical characteristics from the weights.
We observe similar properties in the zoos of \citep{unterthiner_predicting_2020}, see Appendix Section C.
%
Initializing models with random seeds disperses the clusters, compare Figure \ref{fig:dataset_comparison_embeddings} and Table \ref{tab:zoo_comparison}. 
While VAR between 1 fixed and 1 random seed is comparable, VAR$_c$ is considerably smaller with fixed seed, the predictive value of $W$ for the initialization methods drops significantly. 
Random seeds also appear to make both the reconstruction as well as NN property prediction more difficult. 
The repetition of configurations with five seeds hinders shortcuts in the weight space, make reconstruction and the prediction of characteristics harder. 
Thus, we conclude that changing only the seeds results in models with very similar evolution during learning. In these zoos, statistics of the weights perform best. In contrast, using one seed shared across models might create shortcuts in the weight space. Zoos that vary both appear to be most diverse and hardest for learning and NN property prediction. Across all zoos with hyperparameter changes, learned hyper-representations significantly outperform the baselines.
\begin{table*}[t!]
\vspace{-.1in}
\centering
\vspace{.05in}
\begin{minipage}[b]{.99\linewidth}  
\begin{center}
\begin{small}
\begin{sc}
\hspace{-.1in}
{\small
\begin{tabular}{l|p{.55cm}p{.5cm}p{.6cm}|p{.55cm}p{.5cm}p{.6cm}|p{.55cm}p{.5cm}p{.6cm}|p{.55cm}p{.55cm}p{.6cm}}
\toprule
& \multicolumn{3}{c}{\texttt{MNIST-HYP}} & \multicolumn{3}{c}{\texttt{FASHION-HYP}} &  \multicolumn{3}{c}{\texttt{SVHN-HYP}}  & \multicolumn{3}{c}{\texttt{CIFAR10-HYP}}\\
& W & s(W) & E${_{c\text{+}}}$D & W & s(W) & E${_{c\text{+}}}$D & W & s(W) & E${_{c\text{+}}}$D & W & s(W) & E${_{c\text{+}}}$D   \\
\midrule
\texttt{MNIST-HYP} & \cellcolor{darkgray!10} \textbf{.36} & \cellcolor{darkgray!10} .29 & \cellcolor{darkgray!10} \textbf{.36} & .21 & .14 & \textbf{.27} & \textbf{.26} & .12 &   .23 & \text{-}.01 & \text{-}.04 & \textbf{.02} \\
\midrule
\texttt{FASHION-HYP} & \text{-}.02 & \textbf{.08} & .02 & \cellcolor{darkgray!10} .54 & \cellcolor{darkgray!10} .48 & \cellcolor{darkgray!10} \textbf{.56} & .06 &  \textbf{.14} & .01 &  .07 & .10 & \textbf{.27} \\
\midrule
\texttt{SVHN-HYP} & .05 & \textbf{.15} & \text{-}.04 & \text{-}.02 & \textbf{.27} & .10 & \cellcolor{darkgray!10} .44 & \cellcolor{darkgray!10} .34 & \cellcolor{darkgray!10} \textbf{.45} & \text{-}.02 & .08 & \textbf{.10} \\
\midrule
\texttt{CIFAR10-HYP} & \textbf{.11} & .09 & .06 & .38 & .36 & \textbf{.39} & .14 & .14 & \textbf{.15} & \cellcolor{darkgray!10} \textbf{.41} & \cellcolor{darkgray!10} .28 & \cellcolor{darkgray!10} .35 
\end{tabular}
}
\end{sc}
\end{small}
\end{center}
\end{minipage}
\caption{Kendall's $\tau$ score for the generalization gap (GGap) prediction. 
We train estimators for each zoo (rows) and evaluate on all zoos (columns). The block diagonal elements contain the in-distribution prediction values. The remaining values are for out-of-distribution prediction.}
\label{table.out-of-distribution.predictioin}
\vspace{-.15in}
\end{table*}

\textbf{Downstream Tasks}. 
We learn and evaluate our hyper-representations on 11 different zoos: \texttt{TETRIS-SEED}, \texttt{TETRIS-HYP}, 5 variants of \texttt{MNIST}, 
\texttt{MNIST-HYP},  \texttt{FASHION-HYP},  \texttt{CIFAR10-HYP} and \texttt{SVHN-HYP}. We compare to multiple baselines and s(W). The results are shown in Tables \ref{tab:zoo_comparison}, \ref{table.tetris-seed.and.tetris-hyp} and \ref{comparisament.ref}. On all model zoos, hyper-representations learn useful features for the downstream tasks, which outperform the actual NN weights and biases, all of the baseline dimensionality reduction methods as well as s(W) \citep{unterthiner_predicting_2020}.
On the \texttt{TETRIS-SEED},  and \texttt{MNIST-SEED} model zoos, s(W) achieves high $R^2$ scores on all downstream tasks (see Table \ref{table.tetris-seed.and.tetris-hyp} left). 
As discussed above, these zoos contain a strong correlation between $s(W)$ and sample properties.
Nonetheless, learned hyper-representations achieve higher $R^2$ scores on all characteristics, with the exception of \texttt{MNIST-SEED} Eph, where E${_{c\text{+}}}$D is competitive to s(W). 
On \texttt{TETRIS-HYP}, the overall performance of all methods is lower compared to \texttt{TETRIS-SEED} (see Table \ref{table.tetris-seed.and.tetris-hyp} right), making it the more challenging zoo. 
Here, too, hyper-representations have the highest $R^2$ score on all downstream tasks, except for the LR prediction. 
On \texttt{MNIST-HYP},  \texttt{FASHION-HYP},  \texttt{CIFAR10-HYP} and \texttt{SVHN-HYP}, hyper-representations outperform s(W) on all downstream tasks and achieve higher $R^2$ score compared to W on the prediction of continuous hyper-parameters and activation prediction. 
On the remaining categorical hyper-parameters initialization method and optimizer, the weight space achieves the highest $R^2$ scores (see Table \ref{comparisament.ref}). We explain this with the small amount of variation in the model zoo (see Section \ref{sec:zoos_description}), which allows to separate these properties in weight space. 

\textbf{Out-of-Distribution Prediction}. Table \ref{table.out-of-distribution.predictioin} shows the out-of-distribution results for generalization gap prediction\footnote{In Appendix Section D, we also give results for other tasks, including epoch id and test accuracy prediction}, which is a very challenging task. The task is rendered more difficult by the fact that many of the samples which have to be ordered by their performance are very close in performance. 
As the results show, hyper-representations are able to recover the order of samples by performance to a significant degree. Further, hyper-representations outperform the baselines in Kendall's $\tau$ measure in the majority of the results. 
This verifies that our approach indeed preserves the distinctive information about trained NNs while compactly relating to their common properties, including the characteristics of the training data. Presumably, learning representations on zoos with multiple different datasets would improve the out-of-distribution capabilities even further. 
\section{Related Work}

There is ample research evaluating the structures of NNs by visualizing activations, \textit{e.g.}, ~\citep{fleet_visualizing_2014,karpathy_visualizing_2015,yosinski_understanding_2015}, which allow some insights in the patterns of, \textit{e.g.}, the kernels of CNNs. 
Other research evaluated networks by computing a degree of similarity between networks. \citep{laakso_content_2000} compared the activations of NNs by a measure of "sameness". \citep{li_convergent_2015} computed correlations between the activations of different nets.~\citep{wang_towards_2018} tried to match the subspaces of the activation spaces in different networks~\citep{johnson_subspace_2019}, which showed to be unreliable. \citep{raghu_svcca:_2017, morcos_insights_2018, kornblith_similarity_2019} applied correlation metrics to NN activations in order to study the learning dynamics and compare NNs. 
\citep{dinh_sharp_2017} link model properties to characteristics of the loss surface around the minimum solution. 
In contrast comparing models with similarity metrics, other efforts map models to an absolute representation.
~\citep{jia_geometric_2019} approximated the space of DNN activations with a convex hull.~\citep{jiang_predicting_2019} also used activations to approximate the margin distribution and  predict the generalization gap. \citep{corneanu_computing_2020} proposed  persistent homology by using a connectivity patterns in the NN activation, and compute topological summaries. 

While previously mentioned related work studied measures defined on the activations for insights about the NN characteristics, the methods applied on populations of NN weights have not received much attention for the same purpose.
\citep{martin_traditional_2019} relate the empirical spectral density of the weight matrices to accuracy, indicating that the weight space alone contains relevant information about the model. 
\citep{eilertsen_classifying_2020} evalutaed a classifier for hyper-parameter prediction directly from the weights. In contrast, we learn a general purpose hyper-representations in self-supervised fashion.~\citep{unterthiner_predicting_2020} proposed layer-wise statistics derived from the NN models to predict the test accuracy of NN model. 
In \citep{schuerholt2021investigation}, subspaces of the weights space of populations of NNs are investigated for model uniqueness and epoch order. 

In the proposed work, we model associations between the different weights in NN using an attention-based module. This helps us to learn representations that compactly extract the relevant and meaningful information while considering the correlations between the weights in an NN. 
\section{Conclusions}
In this work, we present a novel approach to learn hyper-representations from the weights of neural networks. 
To that end, we proposed novel augmentations, self-supervised learning losses and adapted multi-head attention-based architectures with suitable weight encoding for this domain. Further, we introduced several new model zoos and investigated their properties. We showed that not only learned neural networks but also their hyper-representations contain the latent footprint of their training data. 
We demonstrated high performance on downstream tasks, exceeding existing methods in hyper-parameters, test accuracy, and generalization gap prediction and showing the potential in an out-of-distribution setting. 

\paragraph{Acknowledgements} 
Leading up to this paper, there were many discussions with colleagues, which we would like to acknowledge.
We are particularly grateful to Marco Schreyer, Xavier Giró-i-Nieto, Pol Caselles Rico and Diyar Taskiran.

\paragraph{Funding Disclosure}
This work is supported by the University of St.Gallen Basic Research Fund.

\bibliographystyle{plainnat}
\bibliography{bibliography}

\appendix

\clearpage
\newpage

\addcontentsline{toc}{section}{A. Permutation Augmentation}
\section*{Appendix A. Permutation Augmentation}
In this appendix section, we give the full derivation about the permutation equivalence in the proposed permutation augmentation (Section 2 in the paper).
\label{ap:augmentation}
In the following appendix subsections, we show the equivalence in the \textit{forward} and \textit{backward} pass through the
neural network with original learnable parameters and the permutated neural network.

\addcontentsline{toc}{subsection}{A.1 Neural Networks and Back-propagation}
\subsection*{A.1 Neural Networks and Back-propagation}
Consider a common, fully-connected feed-forward neural network (FFN). It maps inputs $\mathbf{x} \in \mathbb{R}^{N_0}$ to outputs $\mathbf{y} \in \mathbb{R}^{N_L}$. For a FFN with $L$ layers, the forward pass reads as
\begin{equation}
\begin{aligned}  
  \mathbf{a}^0 &= \mathbf{x}, \\
  \mathbf{n}^l &= \mathbf{W}^{l} \mathbf{a}^{l-1} + \mathbf{b}^{l}, \qquad l \in \{1,\cdots,L\}, \\
  \mathbf{a}^l &= \sigma(\mathbf{n}^l), \qquad l \in \{ 1,\cdots,L \}. 
\end{aligned}  
\end{equation}
Here, $\mathbf{W}^l \in \mathbb{R}^{N_{l} \times N_{l-1}}$ is the weight matrix of layer $l$, $\mathbf{b}^l$ the corresponding bias vector. Where $ N_{l}$ denotes the dimension of the layer $l$. 
The activation function is denoted by $ \sigma $, it processes the layer's weighted sum $\mathbf{n}^l$ to the layer's output $\mathbf{a}^l$. 

Training of neural networks is defined as an optimization against a objective function on a given dataset, \textit{i.e.} their weights and biases are chosen to minimize a cost function, usually called \emph{loss}, denoted by $\mathcal{L}$. The training is commonly done using a gradient based rule. Therefore, the update relies on the gradient of $\mathcal{L}$ with respect to weight ${\bf W}^l$ and the bias ${\bf b}^l$, that is it relies on $ \nabla_{\mathbf{W}} \mathcal{L} $ and $ \nabla_{\mathbf{b}} \mathcal{L} $, respectively.
Back-propagation facilitates the computation of these gradients, and makes use of the chain rule to back-propagate the prediction error through the network 
~\citep{rumelhart_learning_1986}. We express the error vector at layer $l$ as
\begin{equation}
  \mathbf{\delta}^l = \nabla_{\mathbf{n}^l} \mathcal{L},
\end{equation}
and further use it to express the gradients as
\begin{equation}
\begin{aligned}  
  \nabla_{\mathbf{W}^l} \mathcal{L} &= \mathbf{\delta}^l (\mathbf{a}^{l-1})^{\mathrm{T}}, \\
  \nabla_{\mathbf{b}^l} \mathcal{L} &= \mathbf{\delta}^l.
\end{aligned}  
\end{equation}

The output layer's error is simply given by
\begin{equation}
  \mathbf{\delta}^L = \nabla_{\mathbf{a}^L} \mathcal{L} \odot \sigma'(\mathbf{n}^L),
\end{equation}
where $\odot$ denotes the Hadamard element-wise product and $\sigma'$ is the activation's derivative with respect to its argument. Subsequent earlier layer's error are computed with
\begin{equation}
\begin{aligned}  
  \mathbf{\delta}^l = (\mathbf{W}^{l+1})^{\mathrm{T}} \mathbf{\delta}^{l+1} \odot \sigma'(\mathbf{n}^l), 
  \qquad l \in  \{1,\cdots,L-1 \}.
\end{aligned}    
\end{equation}
A usual parameter update takes on the form
\begin{equation}
  (\mathbf{W}^l)_{\text{new}} = \mathbf{W}^l - \beta \nabla_{\mathbf{W}^l} \mathcal{L}, \label{ap:eq:update}
\end{equation}
where $\beta$ is a positive learning rate.

\addcontentsline{toc}{subsection}{A.2 Proof: Permutation Equivalence}
\subsection*{A.2 Proof: Permutation Equivalence}

In the following appendix subsection, we show the permutation equivalence for feed-forward and convolutional layers.

\label{sec:headings}
\textbf{Permutation Equivalence for Feed-forward Layers} Consider the permutation matrix $\mathbf{P}^l \in \mathbb{N}^{N_i \times N_l}$, such that $(\mathbf{P}^l)^{\mathrm{T}} \mathbf{P}^l = \mathbf{I}$, where $ \mathbf{I}$ is the identity matrix. 
We can write the weighted sum for layer $l$ as
\begin{equation}
\begin{aligned}  
  \mathbf{n}^{l+1} &= \mathbf{W}^{l+1} \; \mathbf{a}^{l} + \mathbf{b}^{l+1} \\
               &= \mathbf{W}^{l+1} \;\sigma(\mathbf{n}^{l}) +
               \mathbf{b}^{l+1} \\
               &= \mathbf{W}^{l+1} \; \sigma(\mathbf{W}^{l} \mathbf{a}^{l-1} + \mathbf{b}^{l}) + \mathbf{b}^{l+1}.
               \label{data.feed.forward}
\end{aligned}  
\end{equation}
As $\mathbf{P}^l$ is a permutation matrix and since we use the element-wise nonlinearity $\sigma(.)$, it holds that 
\begin{align}
 \mathbf{P}^l \sigma(\mathbf{n}^l) = \sigma(\mathbf{P}^l \mathbf{n}^l),
\end{align}
which implies that we can write
\begin{equation}
\begin{aligned}  
    \mathbf{n}^{l+1}  &= \mathbf{W}^{l+1} \; \mathbf{I} \; \sigma(\mathbf{W}^{l} \; \mathbf{a}^{l-1} + \mathbf{b}^{l}) + \mathbf{b}^{l+1} \\
                &= \mathbf{W}^{l+1} \;(\mathbf{P}^l)^{\mathrm{T}} \; \mathbf{P}^l \; \sigma(\mathbf{W}^{l} \; \mathbf{a}^{l-1} + \mathbf{b}^{l}) + \mathbf{b}^{l+1} \\
                &= \mathbf{W}^{l+1} \; (\mathbf{P}^l)^{\mathrm{T}} \; \sigma( \mathbf{P}^l \; \mathbf{W}^{l} \; \mathbf{a}^{l-1} + \mathbf{P}^l \; \mathbf{b}^{l}) + \mathbf{b}^{l+1} \\
                &= \mathbf{ \hat{W} }^{l+1} \; \sigma( \mathbf{ \hat{W} }^{l} \; \mathbf{a}^{l-1} + \mathbf{ \hat{b} }^{l}) + \mathbf{b}^{l+1}, \label{ap:eq:forward_perm}
\end{aligned}  
\end{equation}
where $ \mathbf{ \hat{W} }^{l+1} = \mathbf{W}^{l+1} (\mathbf{P}^l)^{\mathrm{T}}$, $ \mathbf{ \hat{W} }^{l} = \mathbf{P}^l  \mathbf{W}^{l} $ and $ \mathbf{ \hat{b} }^{l} = \mathbf{P}^l \mathbf{b}^{l} $ are the permuted weight matrices and bias vector.

Note that rows of weight matrix and bias vector of layer $l$ are exchanged together with columns of the weight matrix of layer $l+1$. In turn,  $\forall l \in \{ 1,\cdots,L-1 \}$,  \eqref{ap:eq:forward_perm} holds true. At any layer $l$, there exist $N_l \!$ different permutation matrices $\mathbf{P}^l$. Therefore, in total there are $ \prod_{l=1}^{L-1} \, N_l!$ equivalent networks.

Additionally, we can write
\begin{equation}
\begin{aligned}
(\mathbf{P}^l \mathbf{W}^l)_{\text{new}} =& \mathbf{P}^l\mathbf{W}^l\! - \alpha \mathbf{P}^l \nabla_{\mathbf{W}^l} \mathcal{L} \\ 
\!\!\!\!\!=&  \mathbf{P}^l\mathbf{W}^l\! - \alpha  \mathbf{P}^l \mathbf{\delta}^l (\mathbf{a}^{l-1})^{\mathrm{T}}  \\
\!\!\!\!\!=& \mathbf{P}^l\mathbf{W}^l \!- \alpha  \mathbf{P}^l \left[ (\mathbf{W}^{l+1})^{\mathrm{T}} \mathbf{\delta}^{l+1} \odot \sigma'(\mathbf{n}^l) \right] \, (\mathbf{a}^{l-1})^{\mathrm{T}} \\
\!\!\!\!\!=& \mathbf{P}^l\mathbf{W}^l \!- \alpha  \left[ (\mathbf{W}^{l+1} \mathbf{P}^{\mathrm{T}})^{\mathrm{T}} \mathbf{\delta}^{l+1} \odot \sigma'(\mathbf{P}^l \mathbf{n}^l) \right] \, (\mathbf{a}^{l-1})^{\mathrm{T}} \\
\!\!\!\!\!=& \mathbf{P}^l\mathbf{W}^l \!- \alpha [ (\mathbf{W}^{l+1} (\mathbf{P}^l)^{\mathrm{T}})^{\mathrm{T}} \mathbf{\delta}^{l+1}  \odot  \sigma'( \mathbf{P}^l \mathbf{W}^{l} \mathbf{a}^{l-1}  +  \mathbf{P}^l \mathbf{b}^{l} ) ] \! (\mathbf{a}^{l-1})^{\mathrm{T}}.
\end{aligned}
\end{equation}

If we apply a permutation $\mathbf{P}^l$ to our update rule (equation \ref{ap:eq:update}) at any layer except the last, then using the above, we can express the gradient based update as 
\begin{align}
    &(\mathbf{\hat{W}}^l)_{\text{new}} = \mathbf{\hat{W}}^l - \alpha \left[ (\mathbf{\hat{W}}^{l+1})^{\mathrm{T}} \mathbf{\delta}^{l+1} \odot \sigma'(\mathbf{\hat{W}}^{l} \mathbf{a}^{l-1} +  \mathbf{\hat{b}}^{l} ) \right] \, (\mathbf{a}^{l-1})^{\mathrm{T}} \square \label{ap:eq:update_perm}
\end{align}
The above implies that the permutations not only preserve the structural flow of information in the forward pass, 
but also preserve the structural flow of information during the update with the backward pass. That is we preserve the structural flow of information about the gradients with respect to the parameters during the backward pass trough the feed-forward layers.

\textbf{Permutation Equivalence for Convolutional Layers} We can easily extend the permutation equivalence in the feed-forward layers to convolution layers. 
Consider the 2D-convolution with input channel $\mathbf{x}$ 
and $O$ output channels $\mathbf{a}_{s_1} = [\icol{ \mathbf{a}_1 \\ . \\ \mathbf{a}_O }]$. 
We express a single convolution operation for the input channel ${\bf x}$ with a convolutional kernel ${\bf K}_o, o \in \{1,..., O\}$ as
\begin{equation}
\begin{aligned} 
    \mathbf{a}_o = & \mathbf{b}_o +  \mathbf{K}_o \star \mathbf{x},   o \in \{ 1,\cdots, O \}, \label{permutations.on.kernel} 
\end{aligned} 
\end{equation}
where $\star$ denotes the discrete convolution operation. %

Note that in contrast to the hole set of permutation matrices ${\bf P}^l$ (which where introduced earlier) now we consider only a subset that affects the order of the input channels (if we have multiple) and the order of the concatenation of the output channels. 

We now show that changing the order of input channels does not affect the output if the order of kernels is changed accordingly.

The proof is similar with the permutation equivalence for feed-forward layer. The difference here is that we take into account only the change in the order of channels and kernels. In order to prove permutation equivalence here it suffices to show that we can represent the convolution of multiple input channels by multiple convolution kernels in an alternative form, that is as matrix vector operation.

To do so we fist show that we can express  the convolution of one input channel with one kernel to its equivalent matrix vector product form. Formally, we have that
\begin{equation}
\begin{aligned} 
    \mathbf{a}_o = & \mathbf{b}_o +  \mathbf{K}_o \star \mathbf{x}  =    & \mathbf{b}_o +  \mathbf{R}_o \mathbf{x}, \label{permutations.on.kernel.alternative} 
\end{aligned} 
\end{equation}
where $\mathbf{R}_o$ is the convolution matrix. We build the matrix $\mathbf{R}_o$ in this alternative form \eqref{permutations.on.kernel.alternative} for the convolution operation from the convolutional kernel ${\bf K}_o$. $\mathbf{R}_o$ has a special structure (if we have 1D convolution then it is known as a circulant convolution matrix), while the input channel ${\bf x}$ remains the same. The number of columns in $\mathbf{R}_o$ equals the dimension of the input channel, while the number of rows in $\mathbf{R}_o$ equals the dimension of the output channel. In each row of ${\bf R}_o$, we store the elements of the convolution kernel. That is we sparsely distribute the kernel elements such that the multiplication of one row ${\bf R}_{o,j}$ of ${\bf R}_o$ with the input channel ${\bf x}$ results in the convolution output ${a}_{o,j}$ for the corresponding position $j$ at the output channel ${\bf a}_{o}$. 

The convolution of one input channels by multiple kernels can be expressed as a matrix vector operation. In that case, the matrix in the equivalent form for the convolution with multiple kernels over one input represents a block concatenated matrix, where each of the block matrices has the previously described special structure, \textit{i.e.},
\begin{equation}
\begin{aligned} 
    \mathbf{a}_{s_1} = \left[ \icol{ \mathbf{a}_1 \\ . \\ \mathbf{a}_O } \right] = & \left[ \icol{ \mathbf{b}_{1} \\ . \\ \mathbf{b}_{O} }  \right] +  
    \left[ \begin{matrix}
    \mathbf{R}_1 \\
    . \\
    \mathbf{R}_O
    \end{matrix} \right]  
    \mathbf{x} = 
     \mathbf{b}_{f} +  {\bf R}_{f} \mathbf{x},
    \label{permutations.on.kernel.alternative.block.structure} 
\end{aligned} 
\end{equation}
where $\mathbf{b}_{f} = \left[ \icol{ \mathbf{b}_{1} \\ . \\ \mathbf{b}_{O} }  \right]$ and ${\bf R}_f = \left[ \begin{matrix}
    \mathbf{R}_1 \\
    . \\
    \mathbf{R}_O
    \end{matrix} \right]$.

In the same way the convolution of multiple input channels $\mathbf{x}_{1}, ..., \mathbf{x}_{S}$ by multiple kernels ${\bf K}_1, ..., {\bf K}_O$ can be expressed as a matrix vector operation. In that case, the matrix in the equivalent form for the convolution with multiple kernels over multiple inputs represents a block diagonal matrix, where each of the blocks in the block diagonal matrix has the previously described special structure, \textit{i.e.},
\begin{equation}
\begin{aligned} 
    \left[ \icol{ \mathbf{a}_{s_1} \\ . \\ \mathbf{a}_{s_s}}  \right] = & \left[ \icol{ \mathbf{b}_{f} \\ . \\ \mathbf{b}_{f}}  \right] +  
    \left[ \begin{matrix}
    \mathbf{R}_{f} & {\bf 0} & ... &  {\bf 0} \\
    . &. &. & . \\
    {\bf 0} & ... &  {\bf 0} & \mathbf{R}_{f} \\
    \end{matrix} \right]  
    \left[ \icol{ \mathbf{x}_{1} \\. \\ \mathbf{x}_{S} } \right]=  \left[ \icol{ \mathbf{b}_{f} \\ . \\ \mathbf{b}_{f}}  \right] +  
    {\bf R}  \left[ \icol{ \mathbf{x}_{1} \\. \\ \mathbf{x}_{S} } \right],
    \label{permutations.on.kernel.alternative.block.structure.0.5} 
\end{aligned} 
\end{equation}
where ${\bf R}=\left[ \begin{matrix}
    \mathbf{R}_{f} & {\bf 0} & ... &  {\bf 0} \\
    . &. &. & . \\
    {\bf 0} & ... &  {\bf 0} & \mathbf{R}_{f} \\
    \end{matrix} \right]$, which we can also express as
\begin{equation}
\begin{aligned} 
    \mathbf{a} =  \mathbf{b} +  {\bf R}\left[ \icol{\mathbf{x}_1 \\ . \\ \mathbf{x}_S} \right],
    \label{permutations.on.kernel.alternative.block.structure.1.0} 
\end{aligned} 
\end{equation}
where ${\bf a}=\left[ \icol{ \mathbf{a}_{s_1} \\ . \\ \mathbf{a}_{s_s}}  \right]$ and ${\bf b}=\left[ \icol{ \mathbf{b}_{f} \\ . \\ \mathbf{b}_{f}}  \right]$.

Note that the above equation has equivalent form with equation \eqref{data.feed.forward}, therefore, the previous proof is valid for the update with respect to hole matrix ${\bf R}$. However, ${\bf R}$ has a special structure, therefore for the update of of each element in ${\bf R}$, we can use the chain rule, which results in 
\begin{equation}
\begin{aligned} 
    \frac{\partial f({\bf R})}{\partial R_{ij}}=&\sum_{k}\sum_{l}\frac{\partial f({\bf R})}{\partial R_{kl}}\frac{ \partial R_{kl}}{\partial R_{ij}}= Tr\left[ \left[ \frac{\partial f({\bf R})}{\partial {\bf R}} \right]^T  \frac{\partial {\bf R}}{\partial R_{ij}} \right].
    \label{permutations.on.kernel.alternative.block.structure.element.wise} 
\end{aligned} 
\end{equation}
Replacing $f()$ by $\mathcal{L}$ in the above and using the update rule equation \eqref{ap:eq:update} gives us the update equation for ${R}_{ij}$. Using similar argumentation and derivation that leads to equation \eqref{ap:eq:update_perm} concludes the proof for permutation equivalence for a convolutional layer $\square$

\begin{figure}[t!]
\begin{center}
\begin{minipage}[b]{.49\linewidth}    
    \includegraphics[trim=0in 0in 0in 0in, clip, width=1\linewidth]{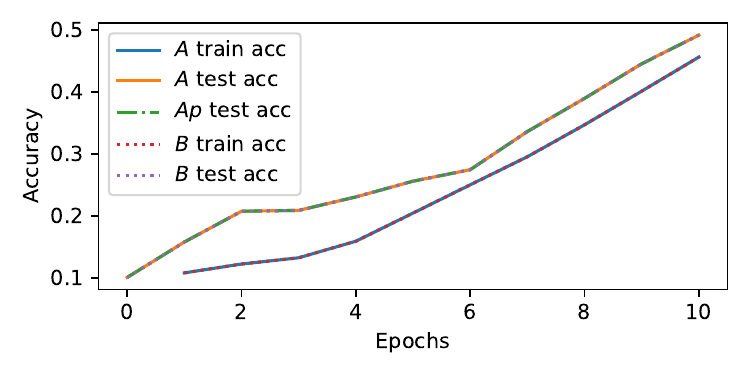}
\end{minipage}        
\begin{minipage}[b]{.49\linewidth}    
    \includegraphics[trim=0in 0in 0in 0in, clip, width=1\linewidth]{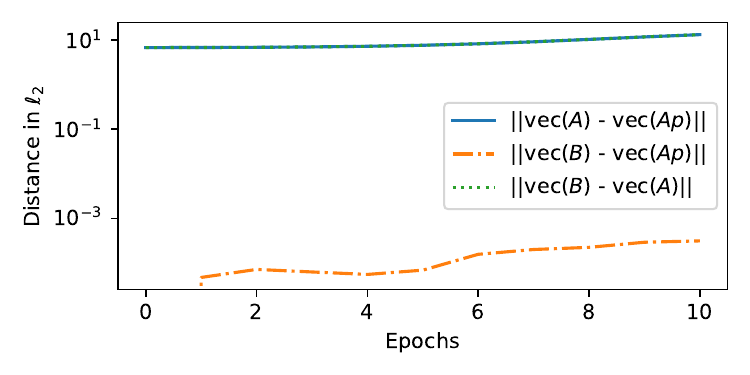}
\end{minipage}        
\caption{Empirical Evaluation of Permutation Equivalence. \textbf{Left:} Accuracies of orignal model $A$, permuted model $Ap$ and model $B$ trained from $Ap$'s initialization. All models are indistinguishable in their accuracies. \textbf{Right:} pairwise distances of vectorized weights over epochs of models $A$, $Ap$ and $B$. The distance between $A$ and $Ap$ as well as $A$ and $B$ is equally large and does not change much over the epochs. The distance between $Ap$ and $B$, which start from the same point in weight space, is small and remains small over the epochs. \textbf{both figures:} permuted versions of models are indistinguishable in their mapping, but far apart in weight space.}
\label{fig:perm_empirical_eval}    
\end{center}
\end{figure}

\textbf{Empirical Evaluation} We empirically confirm the permutation equivalence (see Figure \ref{fig:perm_empirical_eval}). We begin by comparing accuracies of permuted models (Figure \ref{fig:perm_empirical_eval} left). To that end, we randomly initialize model $A$ and train it for 10 epochs. We pick one random permutation, and permute all epochs of model $A$. For the permuted version $Ap$ we compute the test accuracy for all epochs. The test accuracy of model $A$ and $Ap$ lie on top of each other, so the permutation equivalence holds for the forward pass.
To test the backwards pass, we create model $B$ as a copy of $Ap$ at initialization, and train for 10 epochs. Again, train and test epochs of $A$ and $B$ lie on top of each other, which indicates that the equivalence empirically holds for the backwards pass, too.

To track how models develop in weight space, we compute the mutual $\ell_2$-distances between the vectorized weights (Figure \ref{fig:perm_empirical_eval} right). The distance between $A$ and $Ap$, as well as between $A$ and $B$ is high and identical. Therefore, model $A$ is far away from models $Ap$ and $B$. Further, the distance between $Ap$ and $B$ is small, confirming the backwards pass equivalence. We attribute the small difference to numerical errors.

\textbf{Further Weight Space Symmetries}
It is important to note that besides the symmetry used above, other symmetries exist in the model weight space, which change the representation of a NN model, but not it's mapping, \textit{e.g.}, scaling of subsequent layers with piece-wise linear activation functions ~\citep{dinh_sharp_2017}. While some of these symmetries may be used as augmentation, these particular mappings only create equivalent networks in the forward pass, but different gradients and updates in the backward pass when back propagating. Therefore, we did not consider them in our work. 

\clearpage
\newpage

\clearpage
\newpage

\addcontentsline{toc}{section}{B. Downstream Tasks Additional Details}
\section*{Appendix B. Downstream Tasks Additional Details}

In this appendix section, we provide additional details about the downstream tasks which we use to evaluate the utility of the hyper-representations obtained by our self-supervised learning approach.

\addcontentsline{toc}{subsection}{B.1 Downstream Tasks Problem Formulation}
\subsection*{B.1 Downstream Tasks Problem Formulation}
\label{proxy.about.representation.utility}
We use linear probing as a proxy to evaluate the utility of the learned hyper-representations, similar to \cite{grill_bootstrap_2020}. 

We denote the training and testing hyper-representations as ${\bf Z}_{train}$ and ${\bf Z}_{test}$. We assume that training ${\bf t}_{train}$ and testing ${\bf t}_{test}$ target vectors are given. We compute the closed form solution $\hat{\bf r}$ to the regression problem 
\begin{equation*}
    (Q1):\hat{\bf r} = \arg \min_{ {\bf r} } \Vert {\bf Z}_{train}{\bf r}- {\bf t}_{train} \Vert_2^2,
\end{equation*}
and evaluate the utility of ${\bf Z}_{test}$ by measuring the $R^2$ score \cite{wright1921correlation} as discrepancy between the predicted ${\bf Z}_{test}\hat{\bf r}$ and the true test targets ${\bf t}_{test}$. 

Note that by using different targets ${\bf t}_{train}$ in $(Q1)$, we can estimate different ${\bf r}$ coefficients. This enables us to evaluate on different downstream tasks, including, accuracy prediction (Acc), epoch prediction (Eph) as proxy to model versioning, F-Score \cite{goodfellow_deep_2016} prediction (F$_c$), learning rate (LR), $\ell_2$-regularization ($\ell_2$-reg), dropout (Drop) and training data fraction (TF). For these target values, we solve $(Q1)$, but for categorical hyper-parameters prediction, like the activation function (Act), optimizer (Opt), initialization method (Init), we train a linear perceptron by minimizing a cross entropy loss \cite{goodfellow_deep_2016}. Here, instead of $R^2$ score, we measure the prediction accuracy.  

\addcontentsline{toc}{subsection}{B.2 Downstream Tasks Targets}
\subsection*{B.2 Downstream Tasks  Targets}

In this appendix subsection, we give the details about how we build the target vectors in the respective problem formulations for all of the downstream tasks.

\textbf{Accuracy Prediction (Acc)}. In the accuracy prediction problem, we assume that for each trained NN model on a particular data set, we have its accuracy. Regarding the task of accuracy prediction, the value $a_{train,i}$ for the training NN models represents the training target value $t_{train,i}=a_{train,i}$, while the value $a_{test,j}$ for the testing NN models represents the true testing target value $t_{test,j}=a_{test,j}$.

\textbf{Generalization Gap Prediction (GGap)}. In the generalization gap prediction problem, we assume that for each trained NN model on a particular data set, we have its train and test accuracy. The generalization gap represents the target value $g_{i}=a_{train,i}-a_{test,i}$ for the training NN models, while $g_{j}=a_{train,j}-a_{test,j}$ is the true target $t_{test,j}=g_{j}$ for the testing NN models.

\textbf{Epoch Prediction (Eph)}. We consider the simplest setup as a proxy to model versioning, where we try to distinguish between NN weights and biases recorded at different epoch numbers during the training of the NNs in the model zoo. To that end, we assume that we construct the model zoo such that during the training of a NN model, we record its different evolving versions, \textit{i.e.}, the zoo includes versions of one NN model at different epoch numbers $e_{i}$.  
Similarly to the previous task, our targets for the task of epoch prediction are the actual epoch numbers, $t_{train,i}=e_{train, i}$ and $t_{test,j}=e_{test, j}$, respectively.

\textbf{F Score Prediction} (F$_c$). To identify more fine-grained model properties, we therefore consider the class-wise F score. We define the F score prediction task similarly as in the previous downstream task. We assume that for each NN model in the training and testing subset of the model zoo, we have computed F score \cite{goodfellow_deep_2016} for the corresponding class with label $c$ that we denote as F$_{train, c,i}$ and  F$_{test, c, j}$, respectively. Then we use  F$_{train,c,i}$ and F$_{test,c,i}$ as a target value $t_{train,c,i}=$F$_{train,c,i}$ in the regression problem and set  $t_{test,c,j}=$F$_{test,c,j}$ during the test evaluation.

\textbf{Hyper-parameters Prediction}. We define the hyper-parameter prediction task identically as the previous downstream tasks. Where for continuous hyper-parameters, like learning rate (LR), $\ell_2$-regularization ($\ell_2$-reg), dropout (Drop), nonlinear thresholding function (TF), we solve the linear regression problem $(Q1)$. Similarly to the previous task, our targets for the task of hyper-parameters prediction are the actual hyper-parameters values. 

In particular, for learning rate $t_{train,i}=learning\text{ }rate$, for $\ell_2$-regularization $t_{train,i}=\ell_2\text{-}regularization\text{ }type$, for dropout (Drop)  $t_{train,i}=dropout\text{ } value$ and for nonlinear thresholding function (TF) $t_{train,i}=nonlinear\text{ } thresholding\text{ }function$. In a similar fashion, we also define the test targets ${\bf t}_{test}$.

For categorical hyper-parameters, like activation function (Act), optimizer (Opt), initialization method (Init), instead of regression loss $(Q1)$, we train a linear perception by minimizing a cross entropy loss \cite{goodfellow_deep_2016}. Here, also we also define the targets as detailed above. The only difference here is that the targets here have discrete categorical values.

\clearpage
\newpage

\addcontentsline{toc}{section}{C Model Zoos Details}
\section*{Appendix Appendix C. Model Zoos Details}

\begin{table*}[t!]
 \centering
 \begin{minipage}[b]{1\linewidth}
 \centering
 \begin{center}
 \begin{small}
 \begin{sc}
 \begin{tabular}{l|cp{.785cm}p{.785cm}p{2.5cm}p{.5cm}c}
 \toprule
Our Zoos & Data & NN Type & No. Param. &  Varying Prop. & No. Eph & No. NNs  \\
 \midrule
 \texttt{TETRIS-SEED} & TETRIS & MLP & 100 & seed (1-1000) & 75  & 1000*75\\
 \texttt{TETRIS-HYP} & TETRIS & MLP & 100 & seed (1-100), act, init, lr & 75 & 2900*75  \\
 \midrule
 \texttt{MNIST-SEED} & MNIST & CNN & 2464 & seed (1-1000) & 25  & 1000*25\\
  \texttt{FASHION-SEED} & F-MNIST & CNN & 2464 & seed (1-1000) & 25  & 1000*25\\
\midrule
 \texttt{MNIST-HYP-1-FIX-SEED} & MNIST & CNN & 2464 & fixed seed, act, int, lr & 25  & $\sim$ 1152*25\\
 \texttt{MNIST-HYP-1-RAND-SEED} & MNIST & CNN & 2464 & random seed, act, int, lr & 25  & $\sim$ 1152*25\\
 \midrule
 \texttt{MNIST-HYP-5-FIX-SEED} & MNIST & CNN & 2464 & 5 fixed seeds, act, int, lr & 25  & $\sim$ 1280*25 \\
 \texttt{MNIST-HYP-5-RAND-SEED} & MNIST & CNN & 2464 & 5 random seeds, act, int, lr & 25  & $\sim$ 1280*25 \\
 \multicolumn{7}{c}{\text{ }} \\
\multicolumn{7}{c}{\text{ }} \\ 
\toprule
 Existing Zoos  & Data & NN Type & No. Param. & Varying Prop. & No. Eph & No. NNs  \\
 \midrule
 \texttt{MNIST-HYP} & MNIST & CNN & 4970 & act, init, opt, lr, & 9 & $\sim$ 30000*9  \\
 & & &  & $\ell_2$-reg, drop, tf & &  \\
 \texttt{FASHION-HYP} & F-MNIST & CNN & 4970 & act, init, opt, lr, & 9 & $\sim$ 30000*9  \\
 & & &  & $\ell_2$-reg, drop, tf & &  \\
 \texttt{CIFAR10-HYP} & CIFAR10 & CNN & 4970 & act, init, opt, lr,  & 9 & $\sim$ 30000*9  \\
 & & &  & $\ell_2$-reg, drop, tf & &  \\
 \texttt{SVHN-HYP} & SVHN & CNN & 4970 & act, init, opt, lr, & 9 & $\sim$ 30000*9  \\
 & & &  & $\ell_2$-reg, drop, tf & &  
 \end{tabular}
 \end{sc}
 \end{small}
 \end{center}
\end{minipage}
\begin{center}
\vskip .15in
\begin{minipage}[b]{1\linewidth}
\centering
\begin{center}
\begin{small}
\begin{sc}
\begin{tabular}{l|ccccc}
\toprule
&$1$ Init & $M$ Init & No data leakage & Dense check points \\ 
\midrule
\cite{unterthiner_predicting_2020}
& ~~~ $\surd$    & $\surd$ & $\times$ & $\times$ \\ 
\cite{eilertsen_classifying_2020}
& ~~~$\surd$    & $\times$& $\surd$ & $\times$ \\ 
\midrule
proposed zoos & ~~~ $\surd$ & $\surd$ & $\surd$ & $\surd$
\end{tabular}
\end{sc}
\end{small}
\end{center}
\end{minipage}
\end{center}
\caption{Overview of the characteristics for the model zoos proposed and used (existing) in this work.}
 \label{table.intro.characteristics}
\end{table*}

In Table \ref{table.intro.characteristics}, we give an overview of the characteristics for the used model zoos in this paper. This includes
\begin{itemize}
\item The data sets used for zoo creation.
\item The type of the NN models in the zoo.
\item Number of learnable parameters for each of the NNs.
\item Used number of model versions that are taken at the corresponding epochs during training.
\item Total number of NN models contained in the zoo.
\end{itemize}

In Table \ref{table.intro.characteristics} we also compare the existing and the introduced model zoos in prior and this work in terms of properties like initialization, data leakage and presence of dense model versions obtained by recording the NN model during training evolution.

In Table \ref{tab:zoo_arch_details} we provide the architecture configurations and exact modes of variation of our model zoos. 
\begin{table*}[t!]
 \centering
 \begin{minipage}[b]{1\linewidth}
 \centering
 \begin{center}
 \begin{small}
 \begin{sc}
 \begin{tabular}{p{2.0cm}|p{1.8cm}p{1.8cm}p{0.9cm}p{1.3cm}p{0.9cm}p{0.9cm}p{0.7cm}}
 \toprule
Our Zoos & INIT & SEED & OPT &  ACT & LR & DROP & $\ell_2$-Reg  \\
 \midrule
 \texttt{TETRIS-SEED} & Uniform  & 1-1000  & Adam & tanh & 3e-5 & 0.0 & 0.0 \\
 \midrule
 \texttt{TETRIS-HYP} & uniform, normal, kaiming-no, kaiming-un, xavier-no, xavier-un,  & 1-100  & Adam & tanh, relu & 1e-3, 1e-4, 1e-5 & 0.0 & 0.0 \\
 \midrule
 \texttt{MNIST-SEED}  & Uniform  & 1-1000  & Adam & tanh & 3e-4 & 0.0 & 0.0 \\
\midrule
 \texttt{MNIST-HYP- 1-FIX-SEED}  & uniform, normal, kaiming-un, kaiming-no  & 42  & Adam, SGD & tanh, relu, sigmoid, gelu & 3e-3, 1e-3, 3e-4, 1e-4 & 0.0, 0.3, 0.5 & 0, 1e-3, 1e-1 \\
 \midrule
 \texttt{MNIST-HYP- 1-RAND-SEED}   & uniform, normal, kaiming-un, kaiming-no  & 1$\in [1e0, 1e6]$  & Adam, SGD & tanh, relu, sigmoid, gelu & 3e-3, 1e-3, 3e-4, 1e-4 & 0.0, 0.3, 0.5 & 0, 1e-3, 1e-1 \\
 \midrule
 \texttt{MNIST-HYP- 5-FIX-SEED}  & uniform, normal, kaiming-un, kaiming-no  & 1,2,3,4,5  & Adam, SGD & tanh, relu, sigmoid, gelu & 1e-3, 1e-4 & 0.0, 0.5 & 1e-3, 1e-1 \\
 \midrule
 \texttt{MNIST-HYP- 5-RAND-SEED}  & uniform, normal, kaiming-un, kaiming-no  & 5$\in [1e0, 1e6]$  & Adam, SGD & tanh, relu, sigmoid, gelu & 1e-3, 1e-4 & 0.0, 0.5 & 1e-3, 1e-1 \\
 \midrule
  \texttt{FASHION-SEED}  & Uniform  & 1-1000  & Adam & tanh & 3e-4 & 0.0 & 0.0 \\
\toprule
 \end{tabular}
 \end{sc}
 \end{small}
 \end{center}
\end{minipage}
\caption{Architecture configurations and modes of variation of our model zoos.}
\label{tab:zoo_arch_details}
\end{table*}

\begin{figure}[t!]
\begin{center}
\begin{minipage}[b]{.24\linewidth}    
    \includegraphics[trim=0in 0in 0in 0in, clip, width=1\linewidth]{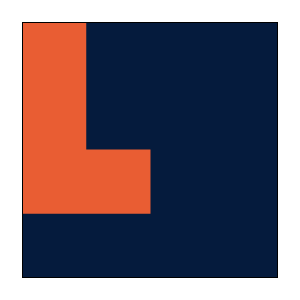}
\end{minipage}        
\begin{minipage}[b]{.24\linewidth}    
    \includegraphics[trim=0in 0in 0in 0in, clip, width=1\linewidth]{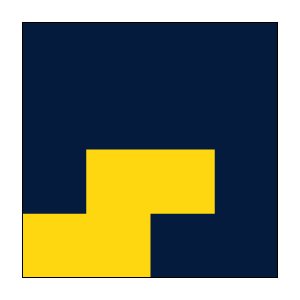}
\end{minipage}        
\begin{minipage}[b]{.24\linewidth}    
    \centering
    \includegraphics[trim=0in 0in 0in 0in, clip, width=1\linewidth]{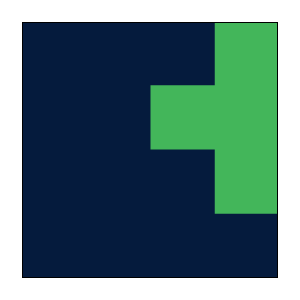}
\end{minipage}        
\begin{minipage}[b]{.24\linewidth}    
    \centering
    \includegraphics[trim=0in 0in 0in 0in, clip, width=1\linewidth]{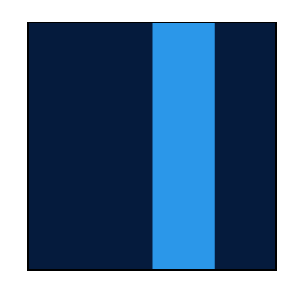}
\end{minipage}
\vskip -0.1in
    \caption{Visualization of samples representing the four basic shapes in our Tetris data set.}
    \label{figure.tetris.shapes}    
\end{center}
\end{figure}

\addcontentsline{toc}{subsection}{C.1 Zoos Generation Using Tetris Data}
\subsection*{C.1 Zoos Generation Using Tetris Data}
\begin{figure*}[t]
\begin{center}
\begin{minipage}[b]{1\linewidth}
\centerline{\includegraphics[trim=0.9in 3.in 5.5in .1in,clip, width=.45\linewidth]{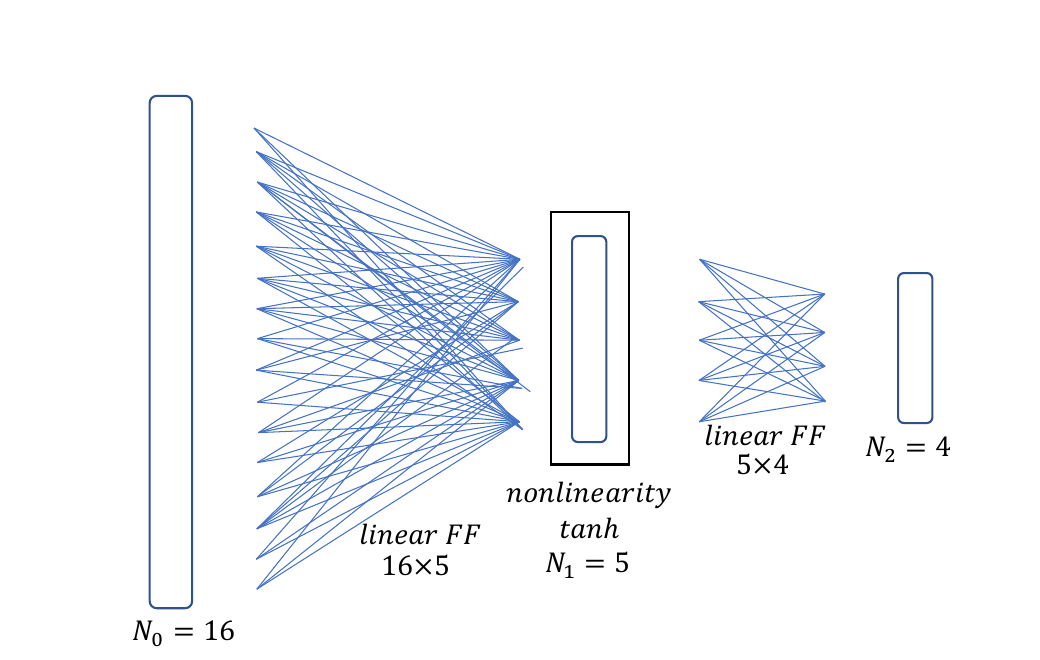}}
\end{minipage}
\caption{A diagram for the feed-forward architecture of the NNs in the  \texttt{TETRIS-SEED} and \texttt{TETRIS-HYP} zoos.}
\label{figure.FNN}
\end{center}
\end{figure*}

As a toy example, we first create a 4x4 grey-scaled image data set that we call \emph{Tetris} by using four tetris shapes. In Figure \ref{figure.tetris.shapes} we illustrate the basic shapes of the tetris data set. We introduce two zoos, which we call \texttt{TETRIS-SEED} and \texttt{TETRIS-HYP}, which we group under \emph{small}. Both zoos contain FFN with two layers. In particular, the FFN has input dimension of $16$ a latent dimension of $5$ and output dimension of $4$. In total the FFN has $16 \times 5 + 5\times 4 =100$ learnable parameters (see Table \ref{tab-ffn-arch}). We give an illustration of the used FFN architecture in Figure \ref{figure.FNN}.

In the \texttt{TETRIS-SEED} zoo, we fix all hyper-parameters and vary only the seed to cover a broad range of the weight space. The \texttt{TETRIS-SEED} zoo contains 1000 models that are trained for 75 epochs. In total, this zoo contains $1000 \times 75 = 75 000$ trained NN weights and biases.

To enrich the diversity of the models, the \texttt{TETRIS-HYP} zoo contains FFNs, which vary in activation function [\texttt{tanh}, \texttt{relu}], the initialization method [\texttt{uniform}, \texttt{normal}, \texttt{kaiming normal}, \texttt{kaiming uniform}, \texttt{xavier normal}, \texttt{xavier uniform}] and the learning rate [$1e$-$3$, $1e$-$4$, $1e$-$5$]. In addition, each combination is trained with 100 different seeds.
Out of the 3600 models in total, we have successfully trained 2900 for 75 epochs - the remainders crashed and are disregarded. So in total, this zoo contains $2900 \times 75 = 217 500$ trained NN weights and biases.

\addcontentsline{toc}{subsection}{C.2 Zoos Generation Using MNIST Data}
\subsection*{C.2 Zoos Generation Using MNIST Data}
\begin{figure*}[t!]
\begin{center}
\begin{minipage}[b]{1\linewidth}
\centerline{\includegraphics[trim=.0in .0in .0in .0in,clip, width=.9\linewidth]{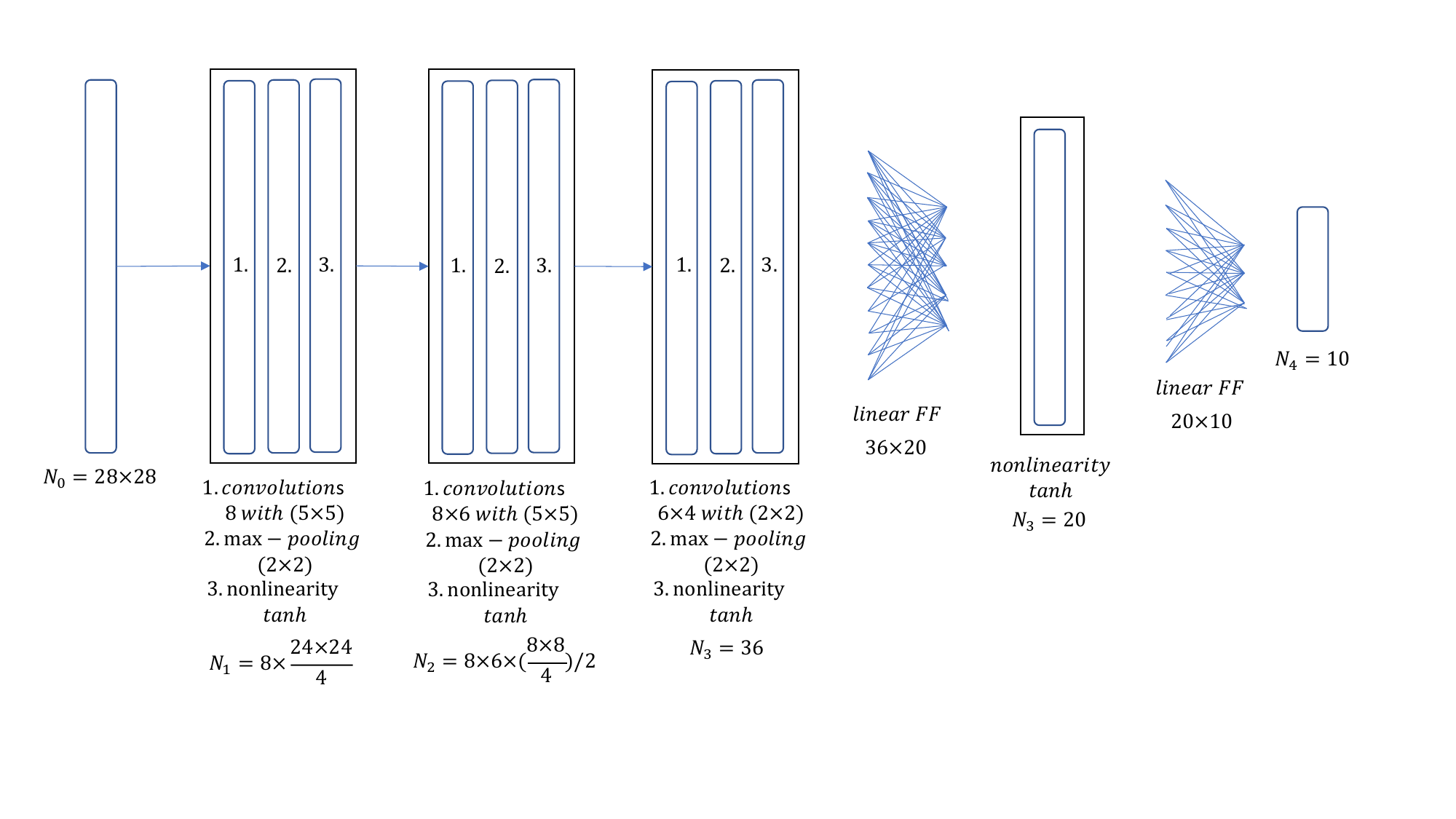}}
\end{minipage}
\caption{A diagram for the CNN architecture of the NNs in the  \texttt{MNIST} zoos.}
\label{figure.CNN}
\end{center}
\end{figure*}

Similarly to \texttt{TETRIS-SEED}, we further create medium sized zoos of CNN models. 
In total the CNN has 2464 learnable parameters, distributed over 3 convolutional and 2 fully connected layers. The full architecture is detailed in Table \ref{tab-cnn-arch}. We give an illustration of the used CNN architecture in Figure \ref{figure.CNN}. Using the MNIST data set, we created five zoos with approximately the same number of CNN models.

In the \texttt{MNIST-SEED} zoo we vary only the random seed (1-1000), while using only one fixed hyper-parameter configuration. In particular, 

In \texttt{MNIST-HYP-1-FIX-SEED} we vary the hyper-parameters. We use only one fixed seed for all the hyper-parameter configurations (similarly to \citep{unterthiner_predicting_2020}). The \texttt{MNIST-HYP-1-RAND-SEED} model zoo contains CNN models, where per each model we draw and use 1 random seeds and different hyper-parameter configuration. 

We generate   \texttt{MNIST-HYP-5-FIX-SEED}  insuring that for each hyper-parameter configurations we add 5 models that share 5 fixed seed. We build \texttt{MNIST-HYP-5-RAND-SEED} such that for each hyper-parameter configurations we add 5 models that have different random seeds.

We 
grouped these model zoos as \emph{medium}. 
In total, each of these zoos approximately $1000 \times 25 = 25 000$ trained NN weights and biases.

\begin{figure*}[t!]
\begin{center}
\begin{minipage}[b]{1\linewidth}
\centerline{\includegraphics[trim=0.5cm 0cm 0.1cm 0cm, clip, width=1\linewidth]{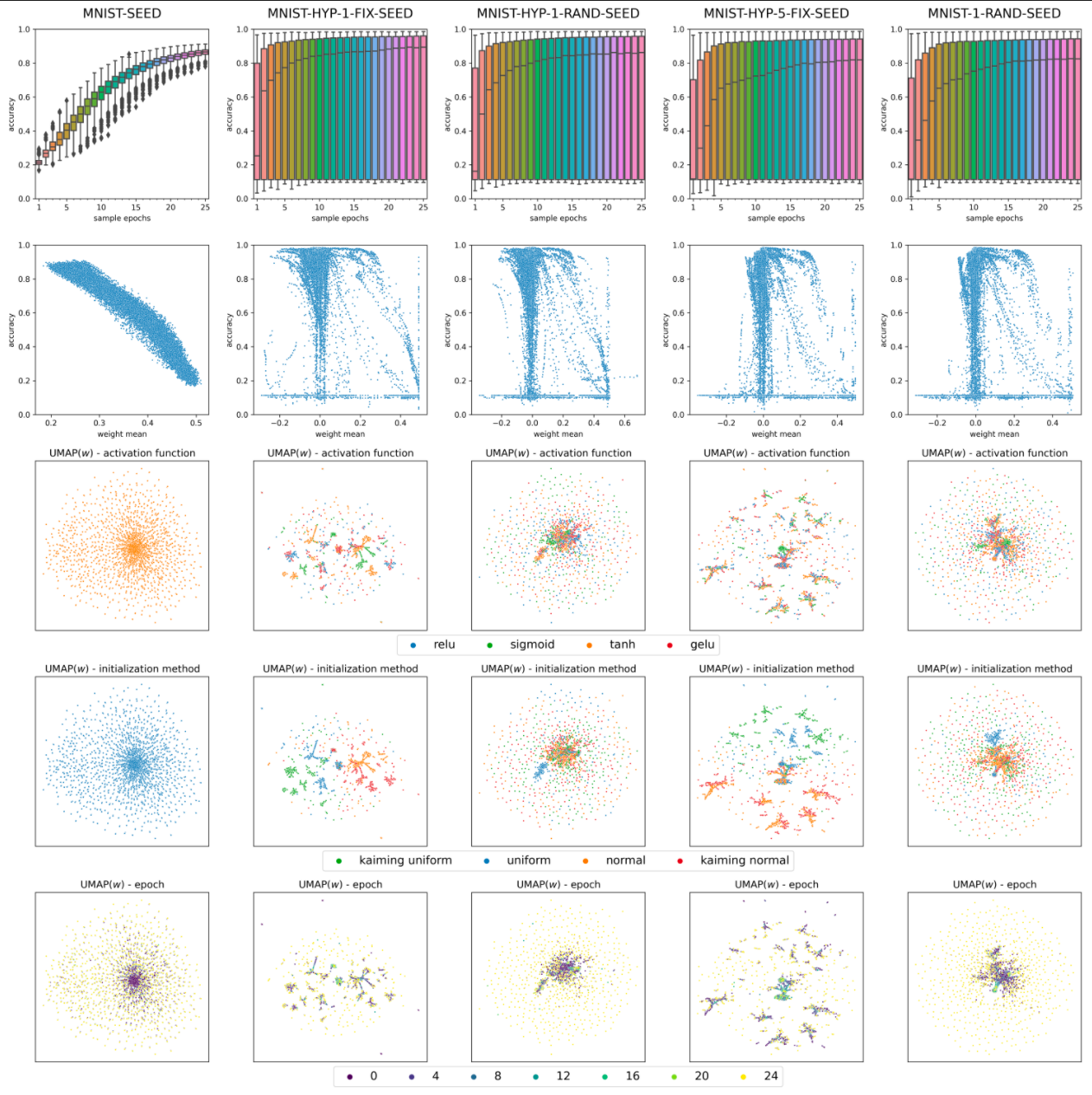}}
\end{minipage}
\caption{Visualization on the properties for the \texttt{MNIST-SEED}, \texttt{MNIST-HYP-1-FIX-SEED}, \texttt{MNIST-HYP-1-RANDOM-SEED}, \texttt{MNIST-HYP-5-FIX-SEED} and \texttt{MNIST-HYP-5-RANDOM-SEED} zoos. \textbf{Row One}. Boxplot of NNs accuracy over the epoch ids. \textbf{Row Two}. NNs accuracy plotted over the mean of the NNs weights of each sample. \texttt{MNIST-SEED} shows homogeneous development and a strong correlation between weight mean and accuracy, while varying the hyperparameters yields heterogeneous development without that correlation. \textbf{Rows Three to Five}. UMAP reductions of the weight space coloured by activation function, initialization method and sample epoch. Zoos with fixed seeds contain visible clusters of NNs that share same initialization method or activation function. Zoos with varying hyperparameters and random seeds do not contain such clear clusters.} 
\label{figure.MNIST.zoo.props}
\end{center}
\end{figure*}

In Figure \ref{figure.MNIST.zoo.props} we provide a visualization for different properties of the \texttt{MNIST-SEED}, \texttt{MNIST-HYP-1-FIX-SEED}, \texttt{MNIST-HYP-1-RANDOM-SEED}, \texttt{MNIST-HYP-5-FIX-SEED} and \texttt{MNIST-HYP-5-RANDOM-SEED} zoos. 
The visualization supports the empirical findings from the paper, that zoos which vary in seed only appear to contain a strong correlation between the mean of the weights and the accuracy. In contrast, the same correlation is considerably lower if the hyper-parameters are varied.
Further, we also observe clusters of models with shared initialization method and activation function for zoos with fixed seeds. Random seeds seem to disperse these clusters to some degree. This additionally confirms our hypothesis about the importance of the generating factors for the zoos. We find that zoos containing hyper-parameters variation and multiple (random) seeds, have a rich set of properties, avoid 'shortcuts' between the weights (or their statistics) and properties, and therefore benefits hyper-representation learning.

\begin{figure*}[t]
\begin{center}
\begin{minipage}[b]{1.00\linewidth}    
    \includegraphics[trim=0in 0in 0in 0in, clip, width=1\linewidth]{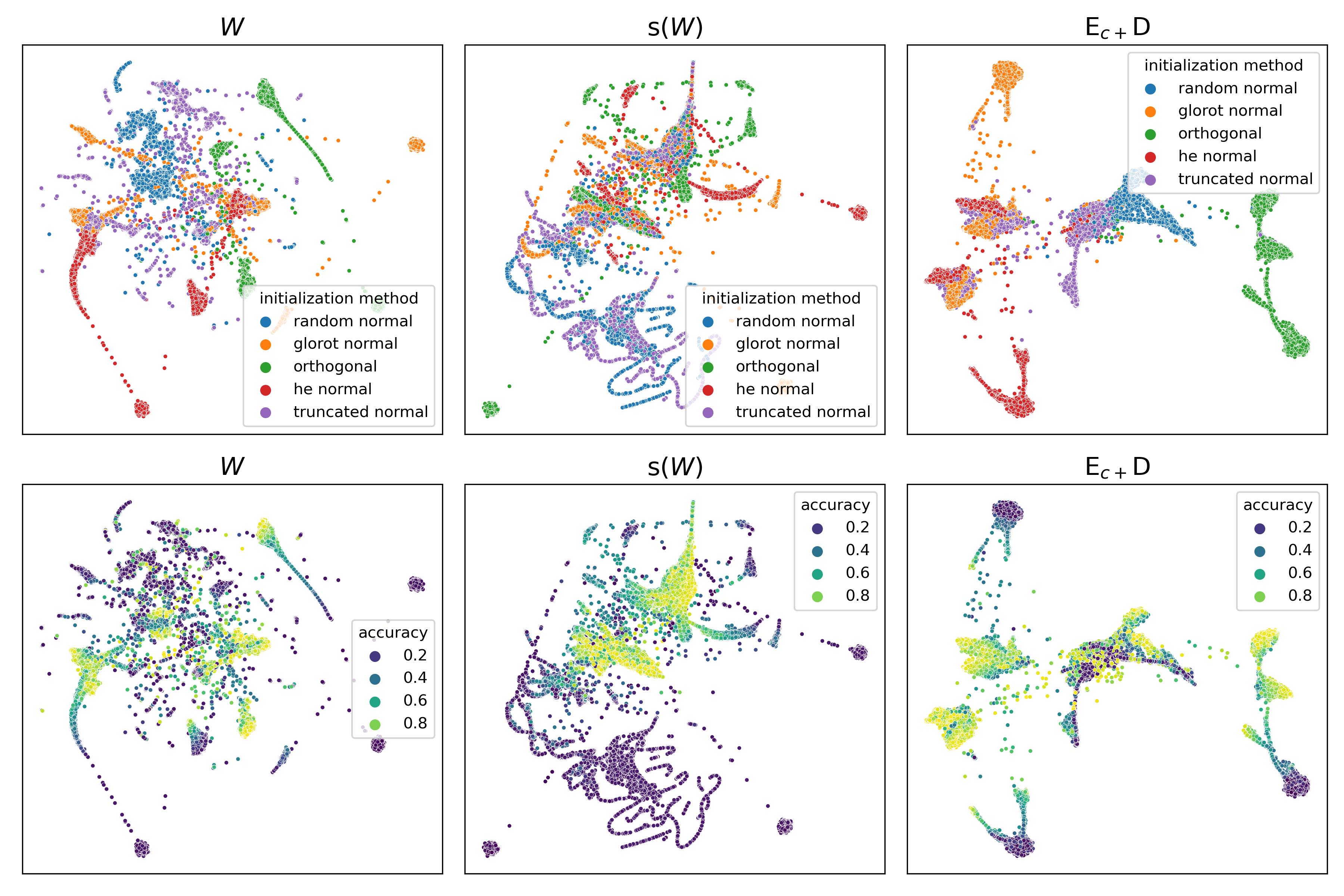}
\end{minipage}        
\vskip -0.1in
\begin{minipage}[b]{1\linewidth}    
\begin{center}
\begin{tabular}{p{3.9cm}p{3.9cm}p{3.9cm}p{3.9cm}}
\end{tabular}
\end{center}
\end{minipage}
\vskip -0.1in
    \caption{UMAP dimensionality reduction of the weight space (left), weight statistics (middle) and learned hyper-representations (right) for the MNIST-HYP zoo \cite{unterthiner_predicting_2020}. 
    The initialization methods for the trained NN weights 
    are already visually separable to a high degree 
    in weight space, which carries over to the learned embedding space, while the statistics introduce a mix between the initialization methods. For accuracy, in seems that the statistics filter out and contain more relevant information than the weight space. Learned embeddings appears to cluster the models according to their initialization methods and within the clusters help to preserve high accuracy.
    }
    \label{fig:embedding_comparsion_mnist_hyper}    
\end{center}
\end{figure*}

In Figure \ref{fig:embedding_comparsion_mnist_hyper}, we show additional UMAP reductions of \texttt{MNIST-HYP}, which confirm our previous findings. Similarly to the UMAP for the \texttt{MNIST-HYP-1-FIX-SEED} zoo, the UMAP for the \texttt{MNIST-HYP} has distinctive and recognizable initialization points. The categorical hyper-parameters are visually separable in weight space. As we can see in the same figure, it seems that the UMAP for the \texttt{MNIST-HYP} zoo contains very few paths along which the evolution during learning of all the models can be tracked in weight space, facilitating both epoch and accuracy prediction.

\addcontentsline{toc}{subsection}{C.3 Zoos Generation Using F-MNIST Data}
\subsection*{C.3 Zoo Generation Using F-MNIST Data}

We used the F-MNIST data set. 
As for the previous zoos for the MNIST data set, we have created one zoos with exactly the same number of CNN models as in \texttt{MNIST-SEED}.
In this zoo that we call \texttt{FASHION-SEED}, we vary only the random seed (1-1000), while using only one fixed hyper-parameter configuration.

\begin{table*}[t!]
\centering
\label{tab-ffn-arch}
\begin{minipage}[b]{1\linewidth}
\begin{center}
\begin{small}
\begin{sc}
\begin{tabular}{l|ccc}
\toprule
 & type & details & Params \\
\midrule
1.1 & Linear    & ch-in=16, ch-out=5 & 80 \\
1.2 & Nonlin. & Tanh \\
\midrule
2.1 & Linear    & ch-in=5, ch-out=4 & 20\\
\end{tabular}
\end{sc}
\end{small}
\end{center}
\caption{FFN Architecture Details. \textsc{ch-in} describes the number of input channels, \textsc{ch-out} the number of output channels.}
\end{minipage}
\centering
\label{tab-cnn-arch}
\vskip 0.1in
\begin{minipage}[b]{1\linewidth}
\begin{center}
\begin{small}
\begin{sc}
\begin{tabular}{l|ccc}
\toprule
 & type & details & Params \\
\midrule
1.1 & Conv    & ch-in=1, ch-out=8, ks=5 & 208 \\
1.2 & MaxPool & ks= 2 \\
1.3 & Nonlin. & Tanh \\
\midrule
2.1 & Conv    & ch-in=8, ch-out=6, ks=5 & 1206 \\
2.2 & MaxPool & ks= 2 \\
2.3 & Nonlin. & Tanh \\
\midrule
3.1 & Conv    & ch-in=6, ch-out=4, ks=2 & 100 \\
3.2 & MaxPool & ks= 2 \\
3.3 & Nonlin. & Tanh \\
\midrule
4 & Flatten & \\
\midrule
5.1 & Linear    & ch-in=36, ch-out=20 & 740 \\
5.2 & Nonlin. & Tanh \\
\midrule
6.1 & Linear    & ch-in=20, ch-out=10 & 200
\end{tabular}
\end{sc}
\end{small}
\end{center}
\caption{CNN Architecture Details. \textsc{ch-in} describes the number of input channels, \textsc{ch-out} the number of output channels. \textsc{ks} denotes the kernel size, kernels are always square.}
\end{minipage}
\end{table*}

\clearpage
\newpage

\addcontentsline{toc}{section}{D. Additional Results}
\section*{Appendix D. Additional Results}

In this appendix section, we provide additional results about the impact of the compression ratio $c=N/L$.

\addcontentsline{toc}{subsection}{D.1 Impact of the Compression Ratio N/L}
\subsection*{D.1 Impact of the Compression Ratio N/L}

In this subsection, we first explain the experiment setup and then comment on the results about the impact of the compression ratio on the performance for downstream tasks.

\textbf{Experiment Setup}. To see the impact of the compression ratio $c=N/L$ on the performance over the downstream tasks, we use our hyper-representation learning approach under different types of architectures, including E${_c}$, ED and E${_c}$D (see Section 3 in the paper). 
As encoders E and decoders D, we used the attention-base modules introduced in Section 3 in the paper.  The attention-based encoder and decoder, on the \texttt{TETRIS-SEED} and \texttt{TETRIS-HYP} zoos, we used 2 attention blocks with 1 attention head each, token dimensions of 128 and FC layers in the attention module of dimension 512. 

We use our weight augmentation methods for representation learning (please see Section 3.1 in the paper). We run our representation learning algorithm for up to 2500 epochs, using the adam optimizer \cite{Kingma:2014:Adam}, a learning rate of 1e-4, weight decay of 1e-9, dropout of 0.1 percent and batch-sizes of 500. In all of our experiments, we use 70\% of the model zoos for training and 15\% for validation and testing each. We use checkpoints of all epochs, but ensure that samples from the same models are either in the train or in the test split of the zoo. As quality metric for the self-supervised learning, we track the reconstruction $R^2$ on the test split of the zoo.

\textbf{Results}. As Table \ref{ablation_latdim} shows, all NN architectures decrease in performance, as the compression ratio increases. 
The purely contrastive setup E$_c$ generally learns embeddings which are useful for the downstream tasks, which are very stable under compression. These results strongly depend on a projection head with enough complexity. The closer the contrastive loss comes to the bottleneck of the encoder, the stronger the downstream tasks suffer under compression.
Notably, the reconstruction of ED is very stable, even under high compression ratios. However, higher compression ratios appear to negatively impact the hyper-representations for the downstream tasks we consider here.
The combination of reconstruction and contrastive loss shows the best performance for $c=2$, but suffers under compression. Higher compression ratios perform comparably on the downstream tasks, but don't manage high reconstruction $R^2$. We interpret this as sign that the combination of losses requires high capacity bottlenecks. If the capacity is insufficient, the two objectives can't be both satisfied.

\begin{table}[t]
\begin{minipage}[b]{1\linewidth}
\centering
\text{\small Encoder with contrastive loss E$_c$}
\label{tab:compression_Ec}
\begin{center}
\begin{small}
\begin{sc}
\begin{tabular}{p{.6cm}|p{.5cm}c p{.6cm}p{.55cm}p{.55cm}p{.55cm}p{.6cm}p{.55cm}}
\toprule
$c$ & Rec & Eph &Acc & Ggap & F$_{C0}$ & F$_{C1}$ & F$_{C2}$ & F$_{C3}$  \\
\midrule
2 & -- & 96.7 & 90.8 & 82.5 & 67.7 & 72.0 & 74.4 & 85.8 \\
3 & -- & 96.6 & 89.4 & 81.5 & 68.4 & 69.4 & 71.1 & 85.1 \\
5 & -- & 96.4 & 89.5 & 81.8 & 67.1 & 68.7 & 69.7 & 84.0 \\
\end{tabular}
\end{sc}
\end{small}
\end{center}
\end{minipage}
\vskip 0.1in
\begin{minipage}[b]{1\linewidth}
\centering
\text{\small Encoder and decoder with reconstruction loss ED }
\label{tab:compression_ED}
\begin{center}
\begin{small}
\begin{sc}
\begin{tabular}{p{.6cm}|p{.5cm}c p{.6cm}p{.55cm}p{.55cm}p{.55cm}p{.6cm}p{.55cm}}
\toprule
$c$ & Rec & Eph &Acc & Ggap & F$_{C0}$ & F$_{C1}$ & F$_{C2}$ & F$_{C3}$  \\
\midrule
2 & 96.1 & 88.3 & 68.9 & 69.9 & 47.8 & 57.2 & 33.0 & 58.1 \\
3 & 93.0 & 74.6 & 69.4 & 66.9 & 53.5 & 46.5 & 38.9 & 48.3 \\
5 & 87.7 & 80.5 & 60.0 & 63.3 & 37.9 & 48.8 & 24.4 & 52.6 \\
\end{tabular}
\end{sc}
\end{small}
\end{center}
\end{minipage}
\vskip 0.1in
\begin{minipage}[b]{1\linewidth}
\centering
\text{\small Encoder and decoder with}
\text{\small reconstruction and contrastive loss E$_{c}$D }
\label{tab:compression_EcD}
\begin{center}
\begin{small}
\begin{sc}
\begin{tabular}{p{.6cm}|p{.5cm}c p{.6cm}p{.55cm}p{.55cm}p{.55cm}p{.6cm}p{.55cm}}
\toprule
$c$ & Rec  & Eph & Acc & Ggap & F$_{C0}$ & F$_{C1}$ & F$_{C2}$ & F$_{C3}$  \\
\midrule
2 & 84.1 & 97.0 & 90.2 & 81.9 & 70.7 & 75.9 & 69.4 & 86.6 \\
3 & 75.6 & 96.3 & 88.3 & 80.7 & 66.9 & 70.8 & 66.1 & 83.2  \\
5 & 64.5 & 96.3 & 85.2 & 80.0 & 63.5 & 68.0 & 61.3 & 73.6 \\
\end{tabular}
\end{sc}
\end{small}
\end{center}
\end{minipage}
\vskip 0.2in
\caption{The impact of the compression ratio $c=N/L$ in the different NN architectures of our approach for learning hyper-representations over the \texttt{Tetris-Seed} Model Zoo. All values are $R^2$ scores and given in \%.}
\label{ablation_latdim}
\end{table}

\addcontentsline{toc}{subsection}{D.2 NN Model Characteristics Prediction on FASHION-SEED}
\subsection*{D.2 NN Model Characteristics Prediction on FASHION-SEED}

Due to space limitations, here in Figure \ref{result.mnist-seed-fashion-seed}, we present the results on the \texttt{FASHION-SEED} together with the results on \texttt{MNIST-SEED}. The experimental setup is same as for the \texttt{MNIST-SEED} zoo, which is explained in the paper. Here, we add a complementary result to our ablation study about the seed variation, that we presented in section 4.3 in the paper. Similarly to the discussion in the paper, random seeds variation in the \texttt{FASHION-SEED} again appears to make the prediction more challenging. The results show that the proposed approach is on-par with the comparing $s(W)$  for this type of model zoos.

\begin{table*}[t!]
\begin{minipage}[b]{1\linewidth}
\begin{center}
\begin{small}
\begin{sc}
\begin{tabular}{l|ccc|ccc}
\toprule
& \multicolumn{3}{c}{\texttt{MNIST-SEED}}& \multicolumn{3}{c}{\texttt{FASHION-SEED} } \\
 & W & s(W) & E${_{c+}}$D & W & s(W) & E${_{c+}}$D  \\
\midrule
Eph &  84.5 & \textbf{97.7} & 97.3 & 87.5 & \textbf{97.0} & 95.8 \\
\midrule
Acc   & 91.3 & 98.7 & \textbf{98.9} & 88.5 & 97.9 & \textbf{98.0} \\
\midrule
GGap   & 56.9 & 66.2 & \textbf{66.7}  & 70.4 & 81.4 & \textbf{83.2} \\

\end{tabular}
\end{sc}
\end{small}
\end{center}
\end{minipage}
\caption{$R^2$ score in \% for epoch, accuracy and generalization gap. 
}
\label{result.mnist-seed-fashion-seed}
\end{table*}

\addcontentsline{toc}{subsection}{D.3 In-distribution and Out-of-distribution Prediction}
\subsection*{D.3 In-distribution and Out-of-distribution Prediction}

\begin{figure*}[t!]
\begin{center}
\begin{minipage}[b]{1\linewidth}
\centerline{\includegraphics[trim=0.7in 0in 0.7in 0in, clip, width=1\linewidth]{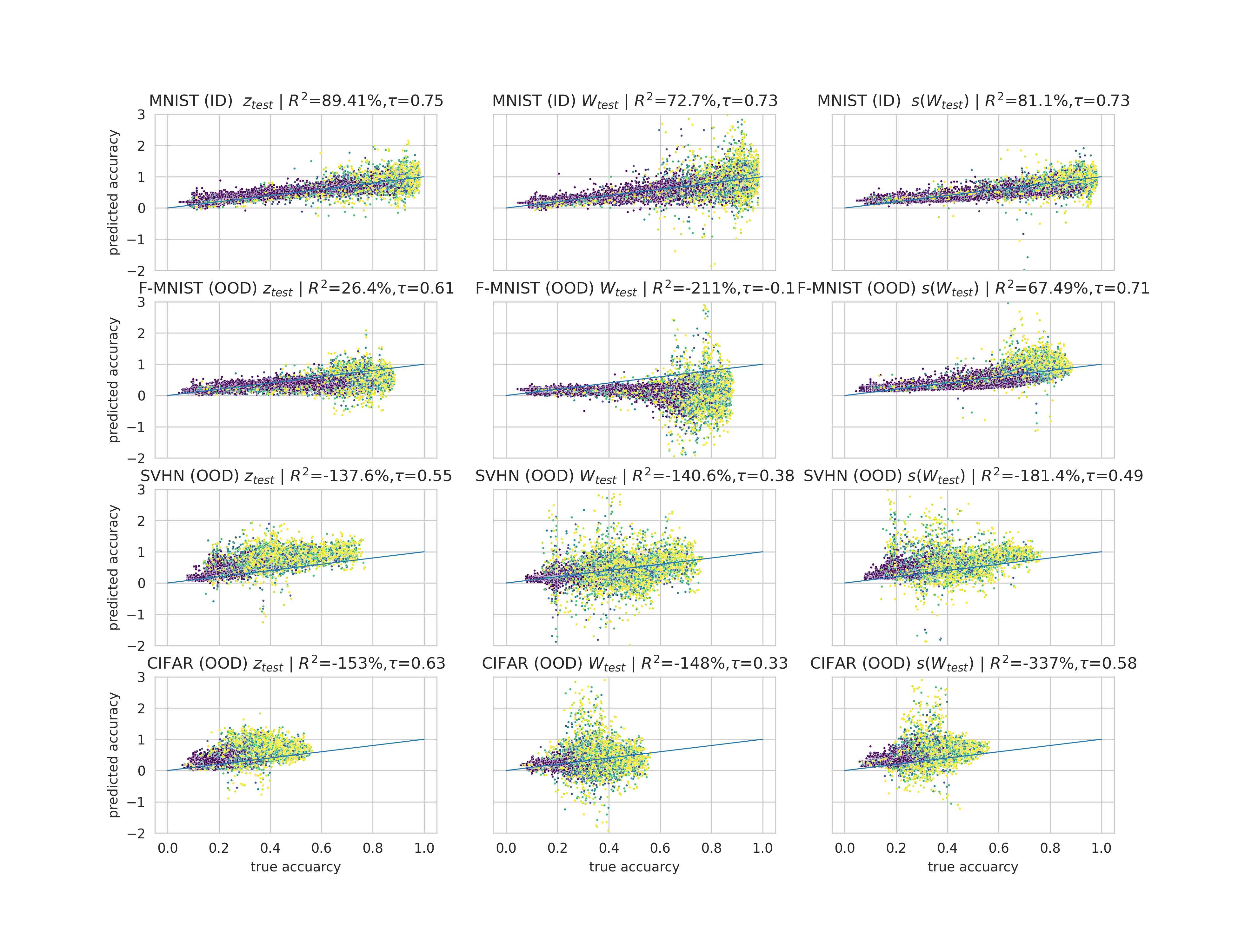}}
\end{minipage}
\begin{minipage}[b]{.99\linewidth}  
\begin{center}
\begin{small}
\begin{sc}
\hspace{-.1in}
{\small
\begin{tabular}{l|p{.55cm}p{.5cm}p{.6cm}|p{.55cm}p{.5cm}p{.6cm}|p{.55cm}p{.5cm}p{.6cm}|p{.55cm}p{.55cm}p{.6cm}}
\toprule
& \multicolumn{3}{c}{\texttt{MNIST-HYP}} & \multicolumn{3}{c}{\texttt{FASHION-HYP}} &  \multicolumn{3}{c}{\texttt{SVHN-HYP}}  & \multicolumn{3}{c}{\texttt{CIFAR10-HYP}}\\
& W & s(W) & E${_{c\text{+}}}$D & W & s(W) & E${_{c\text{+}}}$D & W & s(W) & E${_{c\text{+}}}$D & W & s(W) & E${_{c\text{+}}}$D   \\
\midrule
\texttt{MNIST-HYP} ($\tau$) & \cellcolor{darkgray!10} .73 & \cellcolor{darkgray!10} .73 & \cellcolor{darkgray!10} \textbf{.75} & \text{-}.08 & \textbf{.71} & .61 & .38 & .49 & \textbf{.55} & .33 & .58 & \textbf{.63} \\
\texttt{MNIST-HYP} ($R^2$) & \cellcolor{darkgray!10} 72.7 & \cellcolor{darkgray!10} 81.1 & \cellcolor{darkgray!10} \textbf{89.4} & \text{-}211 & \textbf{67} & 26 & \text{-}140 & \text{-}180 & \text{-}\textbf{137} 
& \text{-}\textbf{148} & \text{-}337 & \text{-}153 \\
\end{tabular}
}
\end{sc}
\end{small}
\end{center}
\end{minipage}
\caption{In-distribution and out-of-distribution results for test accuracy prediction. Representation learning model and linear probes are trained on \texttt{MNIST-HYP}, and evaluated on \texttt{MNIST-HYP}, \texttt{FASHION-HYP}, \texttt{SVHN-HYP} and \texttt{CIFAR-HYP}.}
\label{figure.ood.MNIST.accuracy}
\end{center}
\end{figure*}

\begin{figure*}[t!]
\begin{center}
\begin{minipage}[b]{1\linewidth}
\centerline{\includegraphics[trim=0.7in 0in 0.7in 0in, clip, width=1\linewidth]{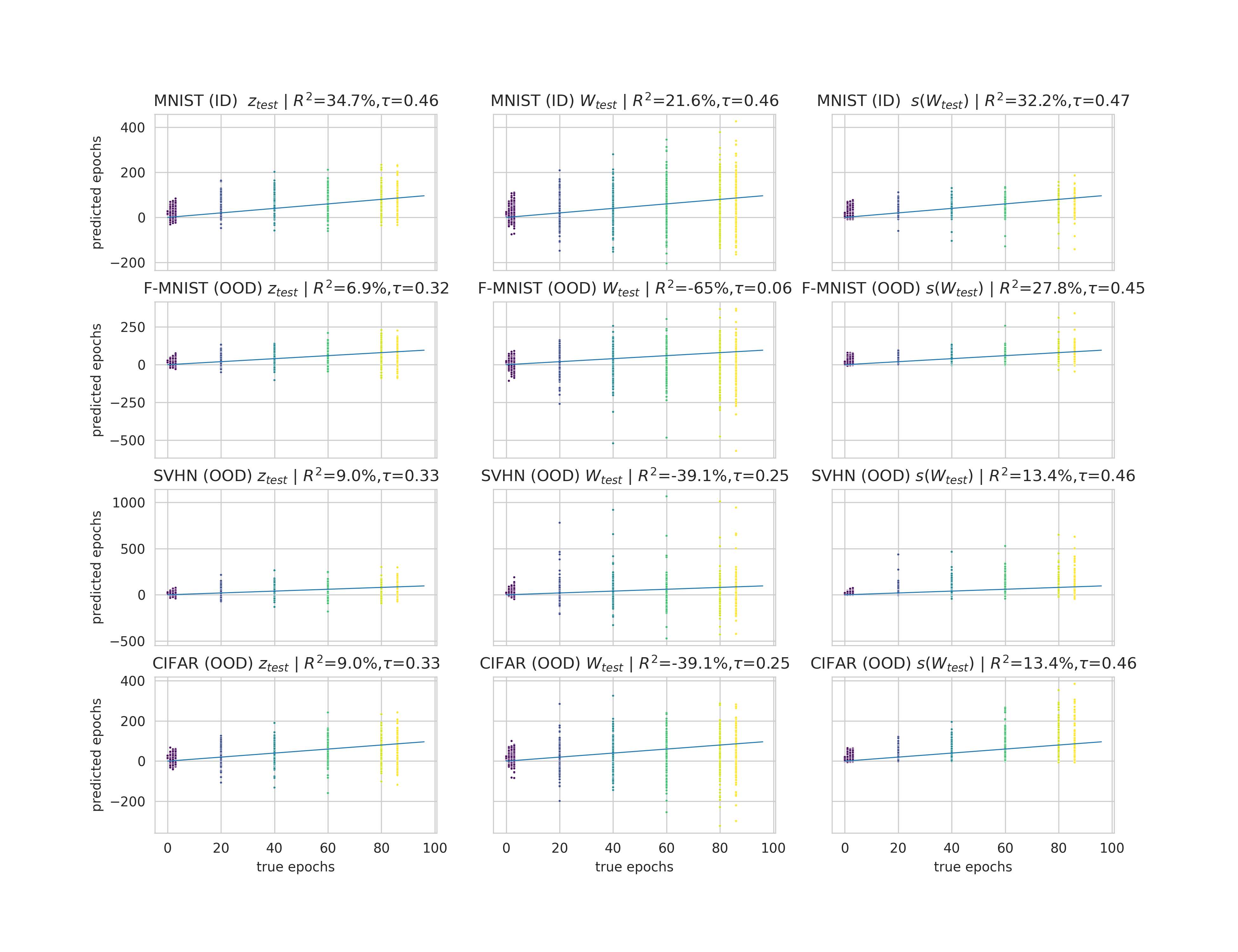}}
\end{minipage}
\vspace{.05in}
\begin{minipage}[b]{.99\linewidth}  
\begin{center}
\begin{small}
\begin{sc}
\hspace{-.1in}
{\small
\begin{tabular}{l|p{.55cm}p{.5cm}p{.6cm}|p{.55cm}p{.5cm}p{.6cm}|p{.55cm}p{.5cm}p{.6cm}|p{.55cm}p{.55cm}p{.6cm}}
\toprule
& \multicolumn{3}{c}{\texttt{MNIST-HYP}} & \multicolumn{3}{c}{\texttt{FASHION-HYP}} &  \multicolumn{3}{c}{\texttt{SVHN-HYP}}  & \multicolumn{3}{c}{\texttt{CIFAR10-HYP}}\\
& W & s(W) & E${_{c\text{+}}}$D & W & s(W) & E${_{c\text{+}}}$D & W & s(W) & E${_{c\text{+}}}$D & W & s(W) & E${_{c\text{+}}}$D   \\
\midrule
\texttt{MNIST-HYP} ($\tau$) & \cellcolor{darkgray!10} .46 & \cellcolor{darkgray!10} \textbf{.47} & \cellcolor{darkgray!10} .46 & .06 & \textbf{.45} & .32 & .25 & \textbf{.46} & .33 & .18 & \textbf{.41} & .16 \\
\texttt{MNIST-HYP} ($R^2$) & \cellcolor{darkgray!10} 21.6 & \cellcolor{darkgray!10} 32.2 & \cellcolor{darkgray!10} \textbf{34.7} & 
\text{-}64.9 & \textbf{27.8} & 6.9 & 
\text{-}39.1 & \textbf{13.4} & 9. & 
\text{-}21.9 & \textbf{19.2} & \text{-}13.\\
\end{tabular}
}
\end{sc}
\end{small}
\end{center}
\end{minipage}
\caption{In-distribution and out-of-distribution results for the epoch id predictions. Representation learning model and linear probes are trained on \texttt{MNIST-HYP}, and evaluated on \texttt{MNIST-HYP}, \texttt{FASHION-HYP}, \texttt{SVHN-HYP} and \texttt{CIFAR-HYP}.}
\label{figure.ood.MNIST.epoch}
\end{center}
\end{figure*}

\begin{figure*}[t!]
\begin{center}
\begin{minipage}[b]{1\linewidth}
\centerline{\includegraphics[trim=0.7in .1in 0.7in 0in, clip, width=1\linewidth]{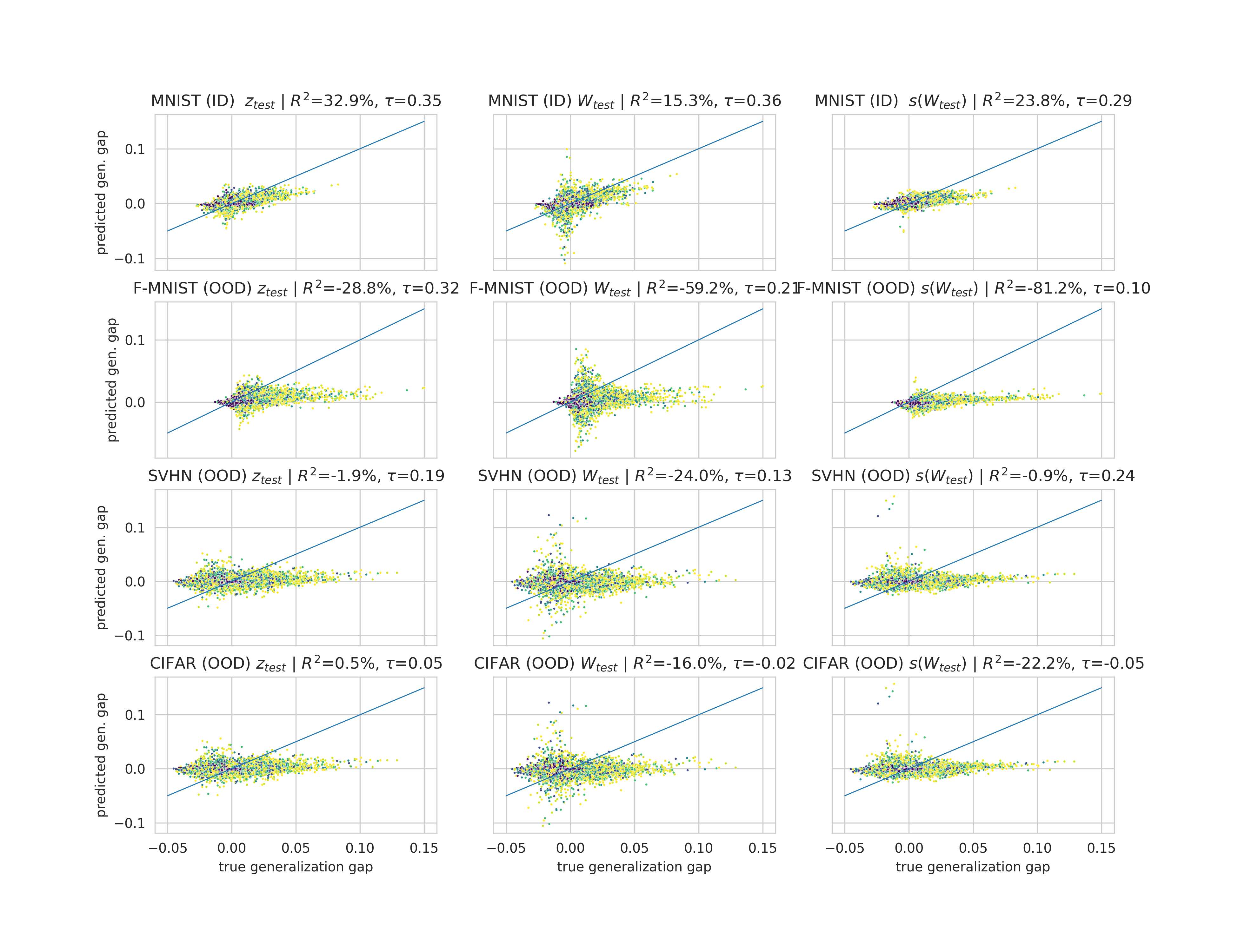}}
\end{minipage}
\begin{minipage}[b]{.99\linewidth}  
\begin{center}
\begin{small}
\begin{sc}
\hspace{-.1in}
{\small
\begin{tabular}{l|p{.55cm}p{.5cm}p{.6cm}|p{.55cm}p{.5cm}p{.6cm}|p{.55cm}p{.5cm}p{.6cm}|p{.55cm}p{.55cm}p{.6cm}}
\toprule
& \multicolumn{3}{c}{\texttt{MNIST-HYP}} & \multicolumn{3}{c}{\texttt{FASHION-HYP}} &  \multicolumn{3}{c}{\texttt{SVHN-HYP}}  & \multicolumn{3}{c}{\texttt{CIFAR10-HYP}}\\
& W & s(W) & E${_{c\text{+}}}$D & W & s(W) & E${_{c\text{+}}}$D & W & s(W) & E${_{c\text{+}}}$D & W & s(W) & E${_{c\text{+}}}$D   \\
\midrule
\texttt{MNIST-HYP} ($\tau$)& \cellcolor{darkgray!10} \textbf{.36} & \cellcolor{darkgray!10} .29 & \cellcolor{darkgray!10} .35 & 
.20 & .10 & \textbf{.32} &
.13 & \textbf{.24} & .19 &
\text{-}.05 & \text{-}.02 & \textbf{.05} \\
\texttt{MNIST-HYP} ($R^2$)& \cellcolor{darkgray!10} 15.3 & \cellcolor{darkgray!10} 24.8 & \cellcolor{darkgray!10} \textbf{32.9} & 
\text{-}56.2 & \text{-}81.8 & \text{-}\textbf{27.8}  
& \text{-}24. & \text{-}\textbf{.9} &
\text{-}1.9 
& \text{-}16. & \text{-}22.2 & \textbf{.5}
\end{tabular}
}
\end{sc}
\end{small}
\end{center}
\end{minipage}
\caption{In distribution and out-of-distribution results for the generalization gap predictions. Representation learning model and linear probes are trained on \texttt{MNIST-HYP}, and evaluated on \texttt{MNIST-HYP}, \texttt{FASHION-HYP}, \texttt{SVHN-HYP} and \texttt{CIFAR-HYP}.}
\label{figure.ood.MNIST.ggap}
\end{center}
\end{figure*}

In Figures \ref{figure.ood.MNIST.accuracy},  
\ref{figure.ood.MNIST.epoch} and
\ref{figure.ood.MNIST.ggap}   
we show in-distribution  and out-of-distribution comparative results for test accuracy, epoch id and generalization gap prediction using the \texttt{MNIST-HYP} zoo.

In the majority of the results for accuracy and generalization gap prediction, our learned representations have higher $R^2$ and Kendall's $\tau$ score. Also, the baseline methods the distribution of predicted target values is more dispersed compared to the true target values. On the epoch id prediction we have comparable results but with lower score, we attribute this to the fact that the zoos contain sparse check points and we suspect that there are not enough so that our learning model could capture the present variability. Overall in the in-distribution and out-of-distribution results for test accuracy, epoch id and generalization gap prediction, the proposed approach has a slight advantage.

\begin{figure*}[t!]
\begin{center}
\begin{minipage}[b]{1\linewidth}
\centerline{\includegraphics[trim=0.0in 0in 0.0in 0in, clip, width=1\linewidth]{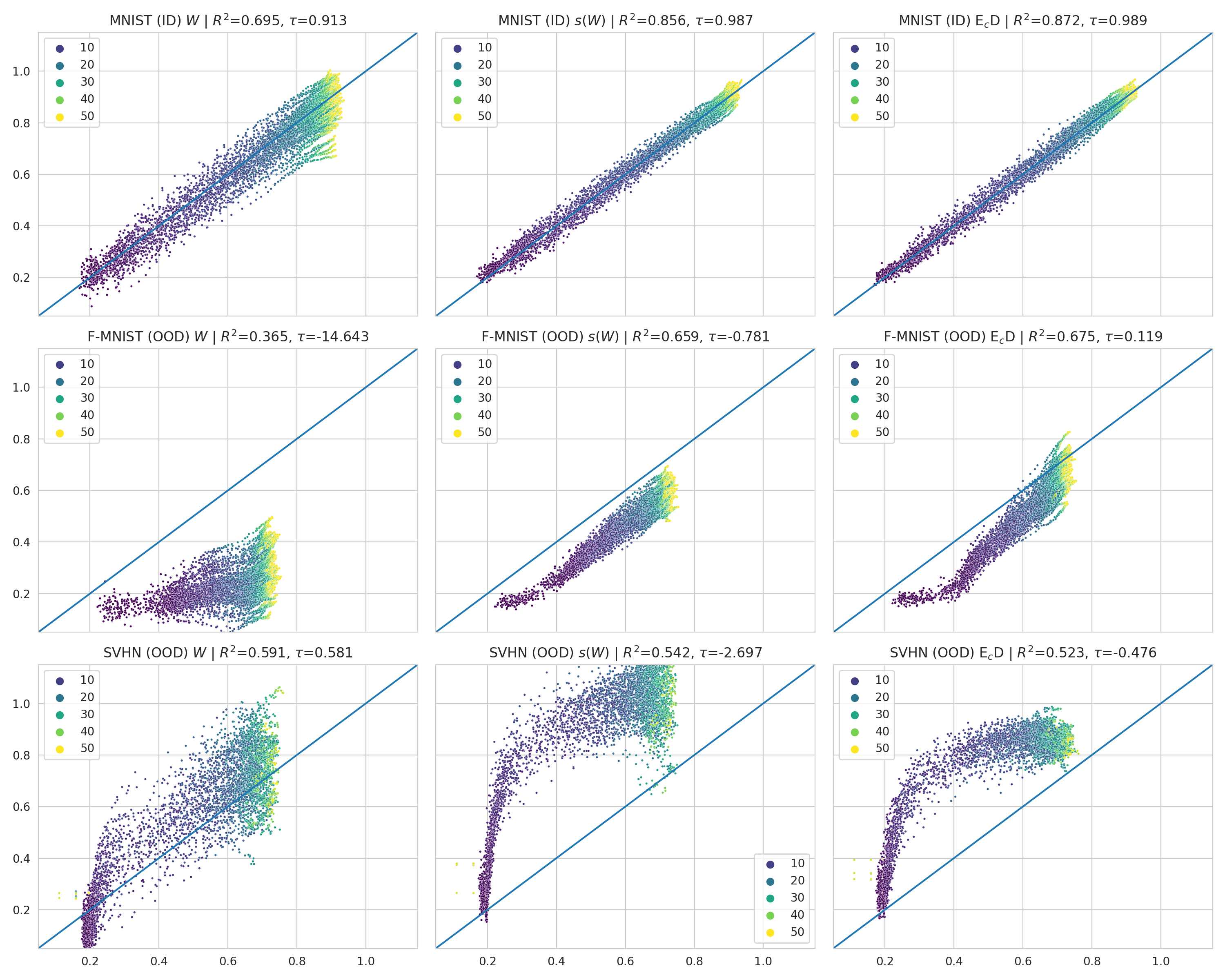}}
\end{minipage}
\begin{minipage}[b]{.99\linewidth}  
\begin{center}
\begin{small}
\begin{sc}
\hspace{-.1in}
{\small
\begin{tabular}{l|p{.75cm}p{.75cm}p{.75cm}|p{.75cm}p{.75cm}p{.75cm}|p{.75cm}p{.75cm}p{.75cm}}
\toprule
& \multicolumn{3}{c}{\texttt{MNIST-SEED}} & \multicolumn{3}{c}{\texttt{FASHION-SEED}} &  \multicolumn{3}{c}{\texttt{SVHN-SEED}}\\
& W & s(W) & E${_{c\text{+}}}$D & W & s(W) & E${_{c\text{+}}}$D & W & s(W) & E${_{c\text{+}}}$D   \\
\midrule
\texttt{MNIST-SEED} ($\tau$) & 
\cellcolor{darkgray!10} .913 & \cellcolor{darkgray!10}.987 & \cellcolor{darkgray!10} \textbf{.989} 
& -14 & -.781 & \textbf{.12} 
& \textbf{.581} & -2.7 & -.48 \\
\texttt{MNIST-SEED} ($R^2$) & 
\cellcolor{darkgray!10} 69.5 & \cellcolor{darkgray!10} 85.6 & \cellcolor{darkgray!10} \textbf{87.2} 
& 36.5 & 65.9 & \textbf{67.5}
& \textbf{.591} & .542 & .523 \\
\\
\end{tabular}
}
\end{sc}
\end{small}
\end{center}
\end{minipage}
\caption{In-distribution and out-of-distribution results for test accuracy prediction. Representation learning model and linear probes are trained on \texttt{MNIST-SEED}, and evaluated on \texttt{MNIST-SEED}, \texttt{FASHION-SEED} and \texttt{SVHN-SEED}.}
\label{figure.ood.MNIST-seed.accuracy}
\end{center}
\end{figure*}

Due to space limitations, for the \texttt{MNIST-SEED}, \texttt{FASHION-SEED} zoos and an additional \texttt{SVHN-SEED} zoo we only include out-of-distribution results for accuracy prediction in Figure \ref{figure.ood.MNIST-seed.accuracy}. Here, too, our learned representations have higher scores in both Kendall's $\tau$ as well as $R^2$. Further, the accuracy prediction for \texttt{SVHN-SEED} clearly preserves the order, but has a noticable bias. We attribute that effect to the different accuracy distributions of \texttt{MNIST-SEED} (ID, accuracy: [0.2,0.95]) and \texttt{SVHN-SEED} (OOD, accuracy: [0.2,0.75]). Due to the higher accuracy in \texttt{MNIST-SEED}, we suspect that the accuracy in \texttt{SVHN-SEED} is overestimated. 


\end{document}